\documentclass[conference]{IEEEtran}
\usepackage{times}

\usepackage[numbers,sort]{natbib}

\usepackage[utf8]{inputenc}             \usepackage[T1]{fontenc}                \usepackage[bookmarks=true]{hyperref}   \usepackage{url}                        \usepackage{booktabs}                   \usepackage{colortbl}                   \usepackage{amsfonts}                   \usepackage{nicefrac}                   \usepackage{microtype}                  \usepackage{multicol}

\usepackage{xcolor}         \usepackage{xcolor-material}

\usepackage{graphicx}

\usepackage{dblfloatfix}

\usepackage{mathtools}
\usepackage{framed}
\usepackage{bbm}
\usepackage{xspace}

\usepackage[short]{optidef}       

\usepackage{amssymb}
\usepackage{amsthm}
\usepackage{amsfonts}
\usepackage{amsmath}
\usepackage{cleveref}

\usepackage{bm}
\usepackage{placeins}

\usepackage{booktabs}
\usepackage{multirow}
\usepackage{tabularray}
\UseTblrLibrary{booktabs}

\usepackage{caption}
\usepackage{etoolbox}

\let\labelindent\relax
\usepackage{enumitem}

\usepackage{caption}
\usepackage{subcaption}

\usepackage{braket}                 \usepackage{xparse}                 

\usepackage[scr=boondox, bb=dsserif]{mathalpha}  \usepackage{bm}

\usepackage{siunitx}
\DeclareSIUnit\ft{ft}               

\usepackage{algorithm}
\usepackage{algpseudocode}
\usepackage{algorithmicx}

\usepackage{nicematrix}             \newcommand\mat[1]{\begin{bNiceMatrix}[light-syntax]
        #1
    \end{bNiceMatrix}}

\usepackage{thmtools}
\declaretheoremstyle[
bodyfont=\normalfont
]{normalstyle}

\usepackage{thm-restate}

\usepackage[framemethod=TikZ]{mdframed}
\mdfdefinestyle{GreyFrame}{nobreak,
    linecolor=white,
    outerlinewidth=0pt,
    roundcorner=2pt,
    leftmargin=-3pt,
    rightmargin=-2pt,
    innertopmargin=2pt, innerbottommargin=2pt, innerrightmargin=3pt,
    innerleftmargin=3pt,
    backgroundcolor=MaterialBlueGrey100!8}

\mdfdefinestyle{ThmFrame}{nobreak,
    linecolor=white,
    outerlinewidth=0pt,
    roundcorner=2pt,
    leftmargin=-3pt,
    rightmargin=-3pt,
    innertopmargin=2pt, innerbottommargin=2pt, innerrightmargin=8pt,
    innerleftmargin=8pt,
    backgroundcolor=MaterialBlueGrey100!15} \definecolor{greentc}{HTML}{52B055}
\definecolor{bluetc}{HTML}{168EF1}
\definecolor{tabblue}{HTML}{1f77b4}
\definecolor{tabred}{HTML}{d62728}

\newcommand{\blue}[1]{{\color{MaterialBlue900}#1}}
\newcommand{\orange}[1]{{\color{MaterialOrange900}#1}}

\newcommand{\mycirc}[1][black]{{\Large\textcolor{#1}{\ensuremath\bullet}}} \newcommand*{\darr}{$\downarrow$}
\newcommand*{\uarr}{$\uparrow$}

\newif\ifarxiv \DeclareFontFamily{U}{BOONDOX-calo}{\skewchar\font=45 }
\DeclareFontShape{U}{BOONDOX-calo}{m}{n}{
  <-> s*[1.05] BOONDOX-r-calo}{}
\DeclareFontShape{U}{BOONDOX-calo}{b}{n}{
  <-> s*[1.05] BOONDOX-b-calo}{}
\DeclareMathAlphabet{\mathcalboondox}{U}{BOONDOX-calo}{m}{n}
\SetMathAlphabet{\mathcalboondox}{bold}{U}{BOONDOX-calo}{b}{n}
\DeclareMathAlphabet{\mathbcalboondox}{U}{BOONDOX-calo}{b}{n}

\DeclareMathOperator{\E}{\mathbb{E}}

\DeclarePairedDelimiterX{\norm}[1]{\lVert}{\rVert}{#1}
\DeclarePairedDelimiterX{\abs}[1]{\lvert}{\rvert}{#1}

\newcommand{\ind}{\mathbb{1}}

\newcommand{\Rb}{\mathbb{R}}

\newcommand{\Eb}{\mathbb{E}}

\newcommand*{\XSet}{\mathcal{X}}
\newcommand*{\USet}{\mathcal{U}}

\newcommand*{\GoalSet}{\mathcal{G}}
\newcommand*{\AvoidSet}{\mathcal{A}}

\newcommand*{\ReachSet}{\mathcal{R}}

\newcommand*{\ZeroSet}{\mathcal{Z}}
\newcommand*{\FiniteSet}{\mathcal{F}}

\DeclareDocumentCommand\vectorbold{ s m }{\IfBooleanTF{#1}{\boldsymbol{#2}}{\mathbf{#2}}} \DeclareDocumentCommand\vb{}{\vectorbold} 

\def\vx{{\vb{x}}}

\let\emptyset\varnothing \def\diffd{\mathrm{d}}

\DeclareDocumentCommand\differential{ o g d() }{ \IfNoValueTF{#2}{
		\IfNoValueTF{#3}
			{\diffd\IfNoValueTF{#1}{}{^{#1}}}
			{\mathinner{\diffd\IfNoValueTF{#1}{}{^{#1}}\argopen(#3\argclose)}}
		}
		{\mathinner{\diffd\IfNoValueTF{#1}{}{^{#1}}#2} \IfNoValueTF{#3}{}{(#3)}}
	}

\DeclareDocumentCommand\derivative{ s o m g d() }
{ \IfBooleanTF{#1}
	{\let\fractype\flatfrac}
	{\let\fractype\frac}
	\IfNoValueTF{#4}
	{
		\IfNoValueTF{#5}
		{\fractype{\diffd \IfNoValueTF{#2}{}{^{#2}}}{\diffd #3\IfNoValueTF{#2}{}{^{#2}}}}
		{\fractype{\diffd \IfNoValueTF{#2}{}{^{#2}}}{\diffd #3\IfNoValueTF{#2}{}{^{#2}}} \argopen(#5\argclose)}
	}
	{\fractype{\diffd \IfNoValueTF{#2}{}{^{#2}} #3}{\diffd #4\IfNoValueTF{#2}{}{^{#2}}}}
}
\DeclareDocumentCommand\dv{}{\derivative}

\makeatletter
\newcommand{\ftype@noticebox}{8}
\newcommand{\notice}{\enlargethispage{2\baselineskip}\@float{noticebox}[b]\footnotesize\@noticestring \end@float }

\newcommand{\@noticestring}{Robotics: Science and Systems 2023
}
\makeatother

\newcommand{\CrefApp}{\Cref}
\newcommand{\eqrefApp}{\eqref}

\arxivtrue 

\begin{document}

\title{Solving Stabilize-Avoid Optimal Control via Epigraph Form and Deep Reinforcement Learning}

\author{Oswin So, Chuchu Fan \\
Massachusetts Institute of Technology \\
\texttt{\{oswinso, chuchu\}@mit.edu} }

\maketitle
\notice

\begin{abstract}
Tasks for autonomous robotic systems commonly require stabilization to a desired region while maintaining safety specifications.
However, solving this multi-objective problem is challenging when the dynamics are nonlinear and high-dimensional, as traditional methods do not scale well and are often limited to specific problem structures.
To address this issue, we propose a novel approach to solve the stabilize-avoid problem via the solution of an infinite-horizon constrained optimal control problem (OCP).
We transform the constrained OCP into epigraph form and obtain a two-stage optimization problem that optimizes over the policy in the inner problem and over an auxiliary variable in the outer problem.
We then propose a new method for this formulation that combines an on-policy deep reinforcement learning algorithm with neural network regression.
Our method yields better stability during training, avoids instabilities caused by saddle-point finding, and is not restricted to specific requirements on the problem structure compared to more traditional methods.
We validate our approach on different benchmark tasks, ranging from low-dimensional toy examples to an F16 fighter jet with a $17$-dimensional state space.
Simulation results show that our approach consistently yields controllers that match or exceed the safety of existing methods while providing ten-fold increases in stability performance from larger regions of attraction.
\ifarxiv
Project page can be found at \url{https://mit-realm.github.io/efppo}.
\fi \end{abstract}

\IEEEpeerreviewmaketitle

\section{Introduction}

Autonomous systems are becoming increasingly prevalent in our lives in recent years; however, designing controllers for complex dynamics systems is difficult. For example, an unmanned aerial vehicle may be required to to observe a target location while maintaining line of sight to a base station \cite{zhang2018cellular}. Another example is the problem of satellite docking that requires approaching a target satellite from a specific direction \cite{jewison2016spacecraft}.
Robot control tasks often involve both stability and safety requirements, where the controller must both drive the system towards and remain stable within some goal region while avoiding unsafe regions. We denote this as the stabilize-avoid problem.
However, synthesizing a policy that achieves both tasks in the presence of input constraints is challenging as these objectives can often be contradictory \cite{jankovic2018robust}.

\vspace{0.05\baselineskip}

\noindent\textbf{Reach-Avoid. }
Reachability analysis and the reach-avoid problem \cite{tomlin2000game} are very closely related to the stabilize-avoid problem that we tackle in this paper. Given a dynamical system, the reach-avoid problem aims to solve for the set of initial conditions and the appropriate control policy to drive a system to a desired goal set while avoiding undesirable states. Hamilton-Jacobi analysis \cite{margellos2011hamilton} provides a methodology for computing the solution to reach-avoid problems,
and is conventionally solved via numerical partial differential equation (PDE) techniques that use state space discretization. These methods are limited in practice to systems with up to $5$ continuous state variables \cite{mitchell2008flexible}. Recent works have applied Deep Reinforcement Learning (DeepRL) to solve reach-avoid problems for higher dimensional systems (e.g., $6$-dimensional system in \cite{hsu2021safety}).
While reaching a goal set is related to stabilization to a goal set, the two objectives have important differences. The goal of the stabilizing controller is to induce stability of the system within a subset of the goal set. This need not be true for the reach controller, where the goal set may not contain any equilibrium point at all. In the worst case, unwanted oscillations could be introduced into the system. We discuss the relationship between the two formulations in \Cref{subsec:reachavoid_relationship}.

\vspace{0.05\baselineskip}

\begin{figure}
    \centering
\includegraphics[width=\linewidth]{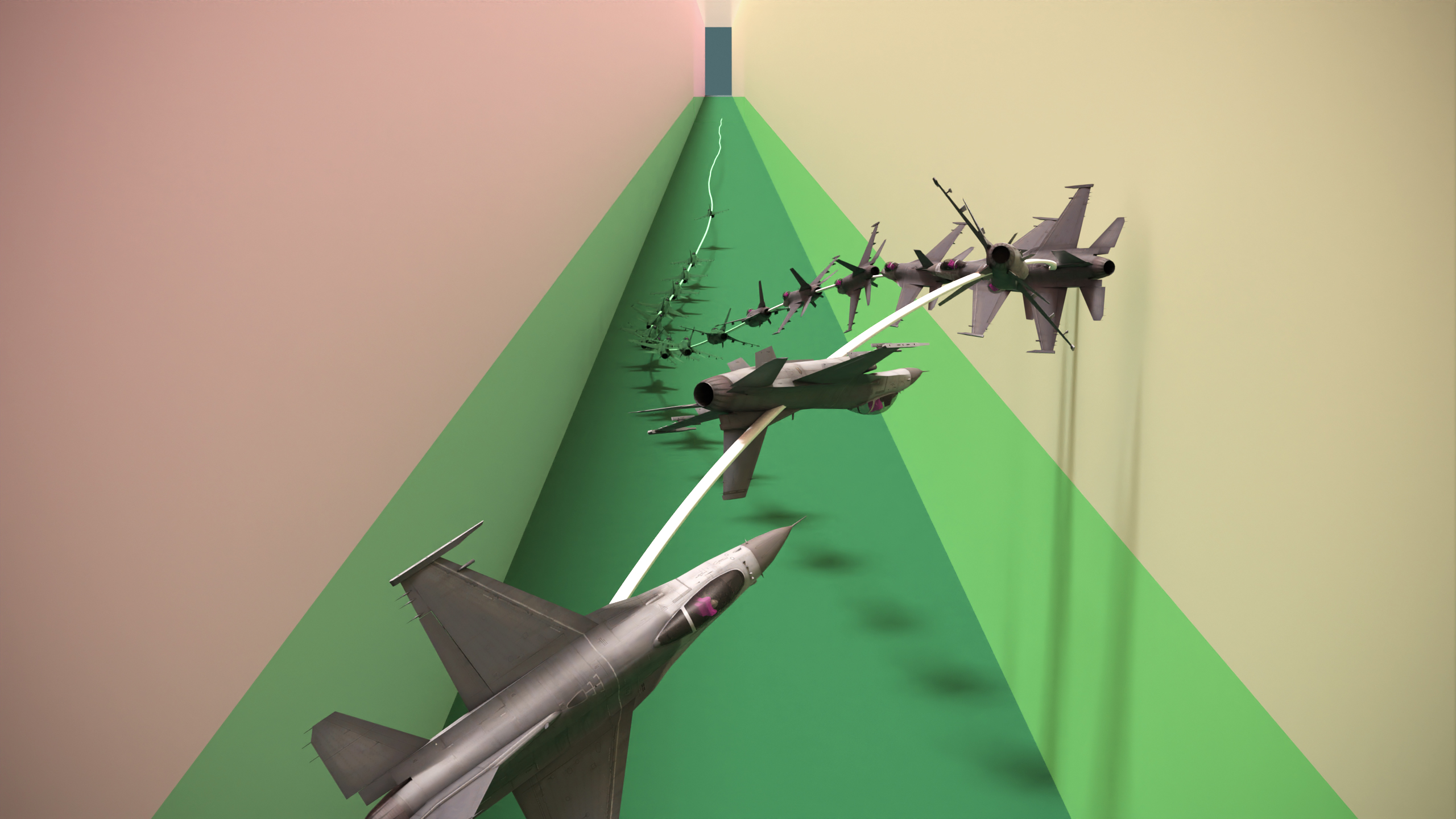}
    \caption{Visualization of F16 ground collision avoidance within a low-altitude flight corridor using EFPPO. The system is required to stabilize to a target altitude near the ground (in \textcolor{MaterialGreen600}{green}), avoid collision with the ground and stay within the flight corridor defined by the two walls and the ceiling.}
    \label{hero:f16}
\end{figure} 
\noindent\textbf{Constrained Reinforcement Learning. }
Works that address the problem of task completion with safety constraints from the reinforcement learning community usually do so from the constrained Markov Decision Process (CMDP) \cite{altman1999constrained} framework. 
Many of these works adopt techniques from constrained optimization to handle the additional safety constraints, of which the use of Lagrangian duality is popular due to its simplicity \cite{achiam2017constrained,tessler2018reward,ma2021feasible,ma2022joint}.
However, the CMDP formulation considers the discounted sum over constraints instead of enforcing the constraint at each state. This allows the additional constraint terms to be treated in the same way as the objective function at the expense of allowing constraint violations. While we can modify the CMDP formulation to disallow all constraint violations, this can lead to an ill-conditioned problems. The relationship between our work and the CMDP setting is discussed in \Cref{subsec:lagr_comparison}.

\vspace{0.05\baselineskip}

\noindent\textbf{Lyapunov Methods. }
Lyapunov theory provides an attractive option for synthesizing safe, stabilizing controllers by using control Lyapunov functions (CLFs) \cite{sontag1983lyapunov,artstein1983stabilization} and control barrier functions (CBFs) \cite{wieland2007constructive,ames2016control}. CLFs and CBFs provide conditions for synthesizing controllers that are certified to be stable and safe respectively. However, they are difficult to construct analytically for general nonlinear systems \cite{giesl2015review}. CLFs / CBFs can be synthesized via convex optimization (e.g., sum-of-squares programming \cite{tan2004searching,ahmadi2016some,dai2022convex}). However, such approaches are limited to systems with polynomial dynamics and rely on the use of solvers for semidefinite programs which can face numerical difficulties \cite{permenter2018partial}. Alternatively, neural networks can be used to synthesize these certificate functions \cite{chang2019neural,qin2021learning,dawson2022safeSurvey}. However, one problem that remains is that CLFs and CBFs cannot be easily combined to yield combined safety and stability guarantees when the set of feasible controls induced by the CLF and CBF do not intersect, forcing the controller to pick one and sacrifice either safety or stability.
This can lead to the presence of unwanted local minima \cite{reis2020control}. Although this can be resolved by learning a joint Control Lyapunov Barrier Function (CLBF) \cite{romdlony2016stabilization,dawson2022safe}, the training process requires the ability to sample from the control-invariant set, which is difficult in the case of complex nonlinear dynamics where the control-invariant set is not known. Our experiments show that this method is difficult to apply in practice. We provide discussions on this in \Cref{sec:discussion}.

\vspace{0.05\baselineskip}

\noindent\textbf{Model Predictive Control. }
Online optimization-based control methods such as model predictive control (MPC) have become increasingly popular for general-purpose control synthesis with the increase in computational power available for robotic systems. Moreover, they can be viewed as a finite-horizon approximation to an infinite horizon optimal control problem (OCP), which is closely linked to Lyapunov stability \cite{grune2017nonlinear}.
However, the constrained OCPs that need to be solved online are computationally expensive, making it difficult to achieve high frequency control updates in practice \cite{rakovic2018handbook}. Moreover, accurate gradient information is typically necessary for solving nonlinear OCPs quickly, making it further difficult to use with dynamics that have expensive gradients or are non-differentiable.
Finally, guaranteeing the recursive feasibility of MPC for general nonlinear systems is challenging \cite{mayne2013apologia}, and in many cases, requires the solution (or approximation) of a control invariant set which can be difficult to find.

\noindent\textbf{Continuous-Time Constrained OCP. }
In addition to the discrete-time formulation used MPC, there exists a number of methods that investigate the problem of constrained OCP in continuous time. Works within this area mainly focus on investigating theoretical properties of the value function for the finite and infinite horizon problems \cite{capuzzo1990hamilton,loreti1987some,soner1986optimal1,soner1986optimal2,mitake2008asymptotic,altarovici2013general}.
We note similarities of our problem formulation to the one discussed in \cite{altarovici2013general}, where the \textit{continuous-time} constrained optimal control problem is solved via transformation into epigraph form. An associated Hamilton-Jacobi PDE and its properties are investigated and used as the basis of a numerical PDE solver for a two-dimensional finite-horizon problem. To the best of our knowledge, the transformation of the infinite-horizon constrained OCP problem into epigraph form for discrete-time problems has not been proposed before in literature.

\vspace{0.1\baselineskip}

In this work, we solve the stabilize-avoid problem by formulating an infinite-horizon constrained OCP,
inspired by the global asymptotic stability guarantees in the unconstrained case \cite{postoyan2016stability}.
Our method for solving the constrained OCP departs from traditional Lagrangian duality based methods and uses an epigraph form which we denote as the Epigraph Form Constrained Optimal Control Problem (EF-COCP). We then solve the EF-COCP using DeepRL by deriving a corresponding policy gradient theorem and applying the proximal policy optimization (PPO) algorithm \cite{schulman2017proximal}. This allows us to tackle a wider range of systems compared to non-RL-based methods and handle general nonlinear non-differentiable black box dynamics with minimal computational cost online.

\noindent\textbf{Contributions. }
We summarize our contributions below.
\begin{itemize}[itemsep=0.2ex]
    \item We propose a new formulation of the safety-constrained OCP via an epigraphic reformulation (EF-COCP) which is easier to interpret and avoids the optimization instability of existing Lagrangian duality methods.
    \item We derive a policy-gradient theorem for the inner problem of EF-COCP and propose the EFPPO algorithm for solving the stabilize-avoid problem using DeepRL.
    \item The proposed EFPPO method is validated on a range of challenging systems, yielding promising empirical results on complex systems such as a $17$-dimensional F16 fighter jet, visualized in \Cref{hero:f16}.
\end{itemize}
 \section{The Stabilize-Avoid Problem}
We consider arbitrary nonlinear discrete-time dynamical systems of the form
\begin{equation} \label{eq:x_dyn}
    x_{k+1} = f(x_k, u_k)
\end{equation}
where $x \in \XSet \subset \Rb^{n_x}$, $u \in \USet \subset \Rb^{n_u}$ and $f : \XSet \times \USet \to \XSet$. In this paper, we consider the following control synthesis problem.
\begin{mdframed}[style=GreyFrame]
\begin{restatable}[Stabilize-Avoid Problem]{problem}{}\label{prob:sa_problem}
    Given a nonlinear system with a goal set $\GoalSet \subset \XSet$ and avoid set $\AvoidSet \subset \XSet$, find a control policy $u = \pi(x)$
that maximizes the size of the set $\ReachSet \subset \XSet$ defined as the set of initial states $x_0$ such that all trajectories started from $x_0$ evolving under the dynamics
    \begin{equation}
        x_{k+1} = f(x_k, \pi(x_k)), \quad x_0 \in \ReachSet,
    \end{equation}
    also satisfy the following two properties.

    \begin{description}[topsep=0.6em,itemsep=0.8em,labelwidth=5em,labelindent=3em]
        \item[\textbf{Stabilize:}] \hspace{2em}$\displaystyle{\limsup_{k \to \infty} \min_{y\in\GoalSet} \norm{x_k - y} = 0}$,
\item[\textbf{Avoid:}] \hspace{2em}$x_k \not \in \AvoidSet$ \; for all $k \geq 0$.
    \end{description}
\end{restatable}
\end{mdframed}
In short, the objective is to reach and (asymptotically) \textit{stabilize} to a goal set $\GoalSet$ while \textit{avoiding} the set of unsafe states $\AvoidSet$. 

\subsection{Relationship with the Reach-Avoid Problem} \label{subsec:reachavoid_relationship}
Note that we use $\limsup$ when defining stabilize instead of minimizing over time as in the reach formulation
\begin{equation} \label{eq:reach_objective}
  \min_k \min_{y \in \GoalSet} \norm{x_k - y} = 0.
\end{equation}
A system that enters but then subsequently exits the goal set $\GoalSet$ will satisfy the reach-avoid problem but not the stabilize-avoid problem, as we illustrate in the following example.
\begin{mdframed}[style=GreyFrame]
\begin{restatable}[Stabilize vs Reach]{example}{}
    Consider a double-integrator with states $\vx = [p, v] \in \Rb^2$, control $u \in \Rb^1$, and the following task specification:
    \begin{center}
    \begin{tblr}{colspec={X[1,c]X[1,c]}}
\textbf{Constraints} & \textbf{Goal} \\
$\abs{u} \leq 1, \;\; \AvoidSet \coloneqq \emptyset$ &
        $\GoalSet \coloneqq \Set{ \vx | p = 1.0 }$
    \end{tblr}
    \end{center}
    One solution to the reach problem (minimizing the first hitting time) yields a controller with periodic orbits, while minimizing the stabilize objective yields a globally stabilizing controller on $\GoalSet$ (\Cref{fig:2dint_reach}).
\end{restatable}
\end{mdframed}

We note that both formulations yield similar results when $\GoalSet$ consists only of equilibrium points.
However, the specifications for $\GoalSet$ may include non-equilibrium points. Applying reach to this problem to obtain stability would require the set of equilibrium points for arbitrary nonlinear dynamics, which is a challenging task in itself to find and may not even exist.
In this work, we consider dynamics for which such a set is not known \textit{a priori}. Hence, applying reach-avoid methods to the stabilize-avoid problem may not give desireable results.
\begin{figure}
    \centering
    \includegraphics[width=0.95\linewidth]{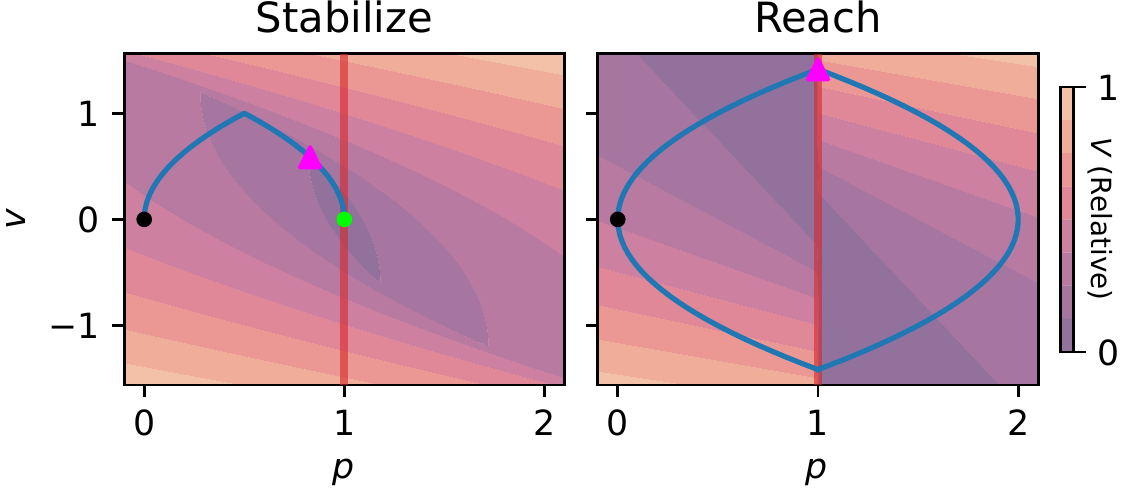}
    \caption{Trajectory and value functions comparing a solution to the stabilize and reach objectives of the double integrator in 1D. The reach controller can reach the goal region at $p=1$ (\textcolor{MaterialRed900}{red line}) faster than stabilize (compare \textcolor{magenta}{magenta triangles}), but the controller never stabilizes to $\GoalSet$ and induces a periodic orbit. The stable controller reaches $\GoalSet$ slower but remains in the set.}
    \label{fig:2dint_reach}
\end{figure} 
\section{Stabilize-Avoid as Infinite-Horizon Constrained Optimal Control Problem}
We tackle \Cref{prob:sa_problem} by solving an infinite-horizon constrained OCP.
To motivate this problem formulation, let $l: \XSet \to \Rb_{\geq 0}$ denote a \textit{non-negative} cost function that takes zero value on $\GoalSet$ and is positive outside $\GoalSet$, and 
define the infinite-horizon \textit{undiscounted} policy value function $V^{l,\pi}$ for an arbitrary policy $\pi$ as
\begin{equation}
    V^{l,\pi}(x_0) \coloneqq \sum_{k=0}^\infty l(x_k), \quad x_{k+1} = f\big(x_k, \pi(x_k) \big).
\end{equation}
Since $l$ is non-negative, $V^{l, \pi}$ is also non-negative.
Using dynamic programming principles, we obtain
\begin{equation}
    V^{l,\pi}(x_k) = l(x_k) + V^{l,\pi}\big( f(x_k, \pi(x_k)) \big).
\end{equation}
The above equations are very close to satisfying the conditions for a discrete-time Lyapunov function \cite{grune2017nonlinear}.
By imposing additional assumptions on $l$, we can show that $V^{l, \pi}$ is a Lyapunov function. We leave the proof in the \CrefApp{app:sec:proof:discr_lyap} for conciseness.
Note that, for a given $\pi$, the set over which stability holds may be very small or even empty. 
This motivates us to ask whether this set can be maximized. The answer is affirmative here: solving the undiscounted infinite-horizon OCP
\begin{mini!}[2]
{\pi}{\sum_{k=0}^\infty l(x_k)}{\label{eq:opt:ocp}}{}
\addConstraint{x_{k+1}}{=f(x_k, \pi(x_k)),}
\end{mini!}
gives a globally asymptotically stabilizing controller under some mild assumptions on the cost function and the controllability of the dynamics \cite{postoyan2016stability}. This provides an answer to \Cref{prob:sa_problem} when safety constraints are not considered.

However, solving an \textit{unconstrained} infinite-horizon OCP does not guarantee satisfaction of the safety constraints $x_k \not \in \AvoidSet$.
Hence, we consider solving a \textit{constrained} infinite-horizon OCP to obtain a policy that is safe by construction.
Let the superlevel set of $h:\XSet \to \Rb$ describes the avoid set
\begin{equation}
    \AvoidSet \coloneqq \Set{ x : h(x) > 0 }.
\end{equation}
We then solve the following \textit{constrained} infinite-horizon OCP
\begin{mini!}[2]
{\pi}{\sum_{k=0}^\infty l(x_k)}{\label{eq:opt:constr_ocp}}{}
\addConstraint{x_{k+1}}{=f(x_k, \pi(x_k))}
\addConstraint{h(x_k)}{\leq 0,\quad}{k \geq 0.}\label{eq:opt:ocp_constr}
\end{mini!}
In the constrained setting, the proof from \cite{postoyan2016stability} that the optimal policy is globally asymptotically stabilizing is not applicable here. However, the proof can be extended to handle this case under certain conditions, which we leave as future work. \section{Safety Constrained Optimal Control via Epigraphic Reformulation}
The previous section describes how solving the stabilize-avoid problem (\Cref{prob:sa_problem}) can be cast as solving an infinite-horizon constrained OCP \eqref{eq:opt:ocp}. In this section, we propose solving the constrained OCP by reformulating the problem into its epigraph form and then solving the resulting two-stage optimization problem.

\subsection{Epigraph Form}
For any optimization problem of the form
\begin{mini!}[2]
{x}{J(x)}{\label{eq:opt:ex:orig}}{}
\addConstraint{h(x)}{\leq 0,}
\end{mini!}
the epigraph form \cite[pp~134]{boyd2004convex} of the above \eqref{eq:opt:ex:orig} is the optimization problem
\begin{mini!}[2]
{x, z}{z}{\label{eq:opt:ex:epi_reform}}{}
\addConstraint{h(x)}{\leq 0} \label{eq:opt:ex:constr1}
\addConstraint{J(x)}{\leq z.} \label{eq:opt:ex:constr2}
\end{mini!}
where $z \in \Rb$ is an auxiliary optimization variable.
It is a standard result in optimization \cite[pp~134]{boyd2004convex} that \eqref{eq:opt:ex:orig} and \eqref{eq:opt:ex:epi_reform} are equivalent.
Now, observe that the constraints \eqref{eq:opt:ex:constr1} and \eqref{eq:opt:ex:constr2} can be combined to yield the following:
\begin{mini!}[2]
{x, z}{z}{\label{eq:opt:ex:linear_obj}}{\label{eq:opt:opt_z_problem}}
\addConstraint{\max\{ h(x), J(x) - z \}}{\leq 0.} \label{eq:opt:ex:join_constr}
\end{mini!}
We can further move the minimization of the $x$ variable into the constraint \eqref{eq:opt:ex:join_constr} (see \CrefApp{app:sec:proof:minimax} for proof) to yield
\begin{mini!}[2]
{z}{z}{\label{eq:opt:ex:linear_obj2}}{\label{eq:opt:opt_z_problem2}}
\addConstraint{\min_x \max\{ h(x), J(x) - z \}}{\leq 0.} \label{eq:opt:ex:join_constr2}
\end{mini!}
This form allows us to convert the original constrained problem \eqref{eq:opt:ex:orig} into an unconstrained \textit{inner problem} over $x$ \eqref{eq:opt:ex:join_constr2} and a \textit{constrained outer} problem \eqref{eq:opt:ex:linear_obj2} over the scalar decision variable $z$.
At the optimal point $(x^*, z^*)$, optimality conditions imply that $z^* = J(x^*)$.
Solving for $z^*$ can thus be thought of as solving for the cost $J$ at the optimal solution. Consequently, if we can bound the value of $J(x^*)$, then $z^*$ will lie within the same bound.
This facilitates treating $z$ as a ``cost budget'' (with units of $J$) for satisfying $h$.
As $z \to \infty$ (i.e., the ``cost budget'' for $J(x)$ increases), $h(x)$ will dominate the $\max$, and $x^*$ will focus on minimizing $h$ more. On the other hand, as $z \to -\infty$ (i.e., the ``cost budget'' for $J(x)$ decreases), $J(x) - z$ will dominate the $\max$, and $x^*$ will focus on minimizing $J$.

\subsection{Epigraph Form Constrained OCP}
We now apply this to the constrained OCP \eqref{eq:opt:constr_ocp}. First, we express the safety constraint \eqref{eq:opt:ocp_constr} equivalently as
\begin{equation}
    \max_{k \geq 0} \,h(x_k) \leq 0.
\end{equation}
Using this, the epigraph form of the constrained OCP \eqref{eq:opt:constr_ocp} reads
\begin{mini!}[2]
{z}{z}{\label{eq:opt:epi1}}{\label{eq:opt:efcocp}}
\addConstraint{\tilde{V}(x_0, z)}{\leq 0,}
\end{mini!}
where the auxiliary value function $\tilde{V}$ is the OCP analogue of the LHS of \eqref{eq:opt:ex:join_constr2}
\begin{mini}[2]
{\pi}{
    \tilde{J}^\pi(x_0, z)\hphantom{acbdefghij}
}{\label{eq:opt:sa_inner}}{\tilde{V}(x_0, z) \coloneqq}
\addConstraint{x_{k+1}}{= f(x_k, \pi(x_k)),}
\end{mini}
with $\tilde{J}^\pi$ defined as
\begin{equation} \label{eq:Jtilde_def}
    \tilde{J}^\pi(x_0, z) \coloneqq \max\left\{ \max_{k \geq 0} h(x_k), \sum_{k=0}^\infty l(x_k) - z \right\}.
\end{equation}
We denote this the epigraph form constrained OCP (EF-COCP).

\subsection{Dynamic Programming for EF-COCP}
Note that \eqref{eq:Jtilde_def} has both a maximization and a sum and hence has a different structure compared to the single sum in the objective function of the typical OCP.
Consequently, the Bellman equation cannot be used in this case.
We derive the corresponding dynamic programming equations below as
\begin{equation} \label{eq:det_dp}
    \tilde{V}(x_k, z_k) = \min_{u_k} \max\Big\{ h(x_k), V\big( x_{k+1}, z_{k+1} \big) \Big\},
\end{equation}
where $z_{k+1}$ has the following ``dynamics''
\begin{equation} \label{eq:z_dyn}
     z_{k+1} = z_k - l(x_k).
\end{equation}
This can again be understood from the intuition of $z$ as a ``cost budget'' in $J$ for satisfying the constraints $h$. Moving from timestep $k$ to $k+1$ incurs the cost $l(x_k)$, which is subtracted from the current ``budget'' $z_k$ to yield the next budget $z_{k+1}$.
If the ``budget'' $z_k$ falls low enough, the cost term in \eqref{eq:Jtilde_def} will dominate the $\max$, and we will have ``run out of budget'' to focus on constraint satisfaction.

\begin{figure*}[ht]
    \centering
    \begin{subfigure}[b]{0.497\linewidth}
        \centering
        \includegraphics[width=\linewidth]{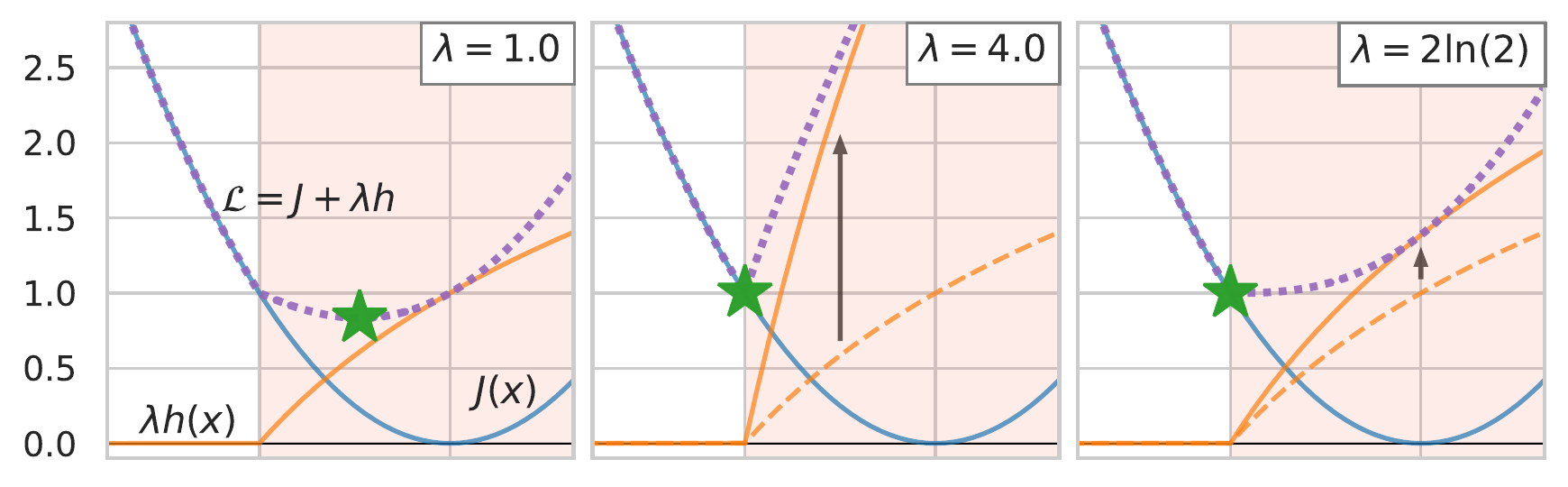}
        \caption{Lagrangian Dual}
        \label{diag:compare_epi:lagr}
    \end{subfigure}\hfill \begin{subfigure}[b]{0.497\linewidth}
        \centering
        \includegraphics[width=\linewidth]{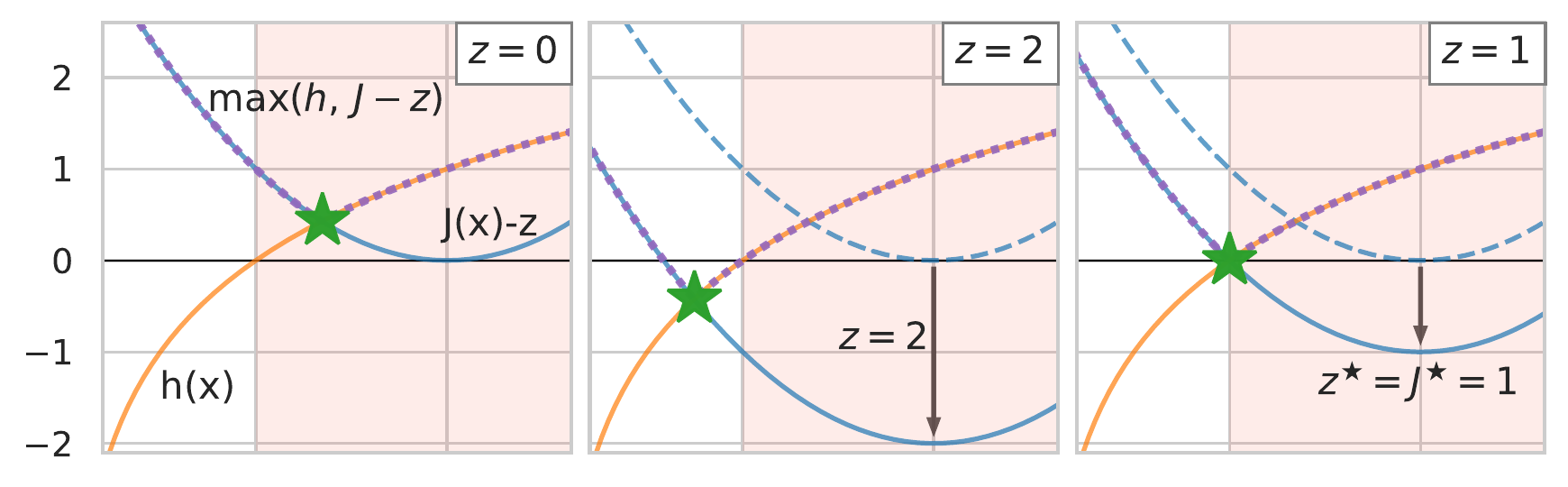}
        \caption{Epigraph Form}
        \label{diag:compare_epi:epi}
    \end{subfigure}
    \caption{Comparison of the inner subproblem for the Lagrangian dual formulation used in CMDP (\textbf{left}) and the epigraphic formulation (\textbf{right}) for the problem $\min_x \blue{J(x)} \textrm{ s.t. } \orange{h(x)} \leq 0$. Note that the gradients of the full objective (\textcolor{MaterialPurple800}{purple}) scale with $\lambda$ (\textbf{left}) but are unaffected in scale by $z$ (\textbf{right}).
}
    \label{diag:compare_epi_lagr}
\end{figure*} 
\subsection{Relationship with Lagrangian Duality} \label{subsec:lagr_comparison}
The epigraph form \eqref{eq:opt:opt_z_problem} shares some similarities with the Lagrangian duality formulation commonly used in (undiscounted) constrained MDPs
\begin{equation} \label{eq:lagr}
    \max_{\lambda \geq 0} \min_{\pi} \underbrace{\sum_{k=0}^\infty l(x_k) + \lambda \sum_{k=0}^\infty [h(x_k)]^+}_{\coloneqq \mathcal{L}(\pi, \lambda)},
\end{equation}
where $[\cdot]^+ = \max(0, \cdot)$, is used to disallow constraints violation.
Both formulations \eqref{eq:lagr}, \eqref{eq:opt:epi1} are two-stage optimization problems, where the outer problem consists of an extra scalar variable ($z, \lambda$ respectively) while the inner problem optimizes with respect to the policy.
The inner problem for both formulations is shown in \Cref{diag:compare_epi_lagr}.
We note the following two differences.

\noindent\textbf{Optimization Stability.}
Due to the $[\cdot]^+$ in \eqref{eq:lagr}, gradients of the inner problem with respect to $\lambda$ will always be non-negative. Consequently, as long as the constraints are not satisfied, $\lambda$ will continue to increase.
However, large values of $\lambda$ are problematic when constraints are not satisfied, since the gradients $\nabla_x \mathcal{L}$ with respect to $x$ scale linearly in $\lambda$. In \cite{stooke2020responsive}, a solution to this problem is proposed by rescaling $\mathcal{L}$ by $1 / (1 + \lambda)$. However, when $\lambda$ is large, the gradients for the $l$ terms will instead vanish. Moreover, since $\lambda$ is a non-decreasing function of the number of optimization iterations, this problem will only become worse as optimization proceeds.

In contrast, since $z$ is additive within the $\max$, the scale of gradients is not altered. Consequently, EF-COCP does not suffer from this issue of optimization instability.

\noindent\textbf{Intuition.}
The auxiliary variable $z$ in the epigraph form is in units of cost and represents a cost budget, as shown earlier. On the other hand, the Lagrange multiplier $\lambda$ is a ratio representing the cost per unit constraint, but this is harder to interpret when cost and constraints cannot be easily compared. Consequently, it is much easier to estimate upper bounds for $z$. We take advantage of this to bound the range of $z$ used for solving the inner problem, which we discuss in the next section. \section{Solving EF-COCP with Deep Reinforcement Learning}
The previous section introduces a new epigraph form of the constrained optimal control problem, but does not provide a method of solving this formulation. In this section, we tackle this problem via reinforcement learning and introduce a framework for learning controllers for complex, nonlinear, potentially non-smooth dynamics.

Given that the inner optimization problem of the epigraph form \eqref{eq:opt:sa_inner} still retains many similarities with the original problem,
we choose to solve for the policy and the value function using reinforcement learning.
Specifically, we use Proximal Policy Optimization (PPO) \cite{schulman2017proximal} but with modified definitions of the value function, advantage functions, returns, and generalized advantage estimator (GAE) \cite{schulman2015high}.

\begin{figure*}[t!]
    \centering
    \includegraphics[width=0.95\linewidth]{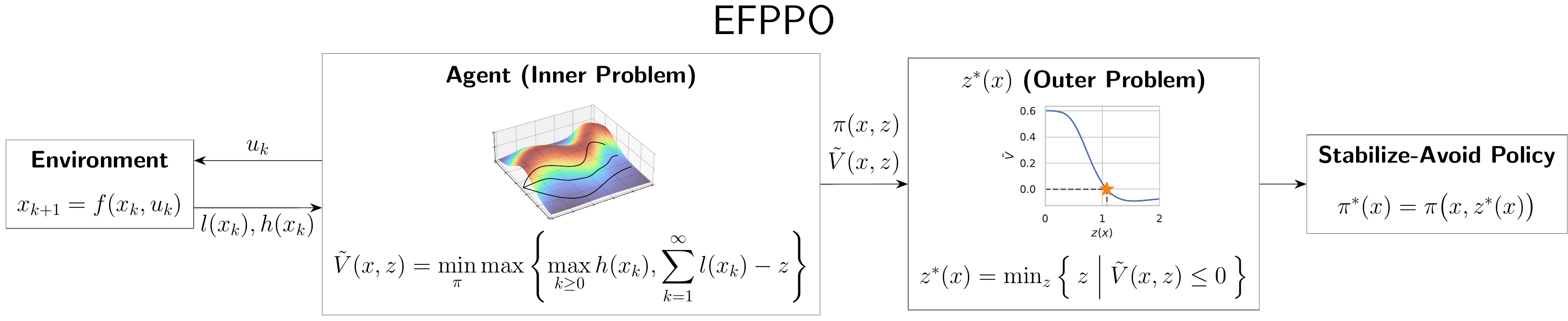}
    \caption{Summary of the EFPPO algorithm. First, reinforcement learning is used to solve the inner problem \eqref{eq:opt:sa_inner} and learn $\tilde{V}(x, z)$ and $\pi(x, z)$ over the entire state space. Then, the optimal $z^*$ which solves the outer problem \eqref{eq:opt:epi1} is regressed.}
    \label{diag:efppo}
\end{figure*} 
While we have treated the policy $\pi$ as a deterministic function, we will use a stochastic $\pi$ for the purpose of improved exploration while performing Deep RL, where $\pi(u|x)$ now defines a distribution over controls. However, while $\E[a+b] = \E[a] + \E[b]$, the same does not hold over the $\max$ operator used to define $\tilde{J }^{\pi}$ in \eqref{eq:Jtilde_def}. Consequently, we need to be careful when defining $\tilde{V}^{\pi}$ for a \textit{stochastic} policy such that an analogous dynamic programming equation to the deterministic case \eqref{eq:det_dp} can be applied. Consider the following \textit{nested expectation} form of an OCP policy value function:
\begin{subequations}
\begin{align}
    &\mathrel{\phantom{=}} V^{\pi}(x_0) \notag \\
    &= \lim_{K \to \infty} \Eb_{0:K}\!\left[\sum_{k=0}^{K} l_k \right], \\
    &= \lim_{K \to \infty} \Eb_{0}\!\bigg[l_0 + \dots + \Eb_{K-1}\Big[l_{K-1} + \Eb_{K}[l_K] \Big] \bigg], \label{eq:stoch_V_sum:nested} \\ 
    &= \Eb_{0}[l_0 + V^{\pi}(x_1)], \label{eq:stoch_V_sum:dp}
\end{align}
\end{subequations}
where, for conciseness, we denote
\begin{equation}
    \Eb_k \coloneqq \Eb_{u_k | x_k}, \quad \Eb_{k:t} \coloneqq \Eb_{u_k, \dots u_t | x_k}, \quad l_k \coloneqq l(x_k).
\end{equation}
Note how the nested expectations of \eqref{eq:stoch_V_sum:nested} lends itself to the dynamic programming equations of \eqref{eq:stoch_V_sum:dp}. In the case of EF\nobreakdash-COCP, we define $\tilde{V}^\pi$ analogously to obtain
\begin{subequations} \label{eq:stoch_V_efocp:def}
\begin{align}
    &\mathrel{\phantom{=}} \tilde{V}^{\pi}(x_0, z_0) \\
    &= \!\lim_{K \to \infty}\! \Eb_{0}\!\bigg[h_0 \vee \dots \vee \Eb_{K-1}\Big[h_{K-1} \vee \Eb_{K}[h_K \vee \sum_{k=0}^K l_k - z_0] \Big] \bigg], \label{eq:stoch_V_efocp:nested} \\ 
    &= \Eb_{0}\big[h_0 \vee \tilde{V}^{\pi}(x_1, z_1) \big], \label{eq:stoch_V_efocp:dp}
\end{align}
\end{subequations}
where we have used $a \vee b \coloneqq \max(a, b)$ for conciseness and where $x_k, z_k$ follow the dynamics \eqref{eq:x_dyn} and \eqref{eq:z_dyn}.
We also define the action-value function $\tilde{Q}^{\pi} : \XSet \times \Rb \times \USet \to \Rb_{\geq 0}$ as 
\begin{equation}
    \tilde{Q}^{\pi}(x_k, z_k, u_k) = \max\Big( h(x_k), \; \tilde{V}^{\pi}(x_{k+1}, z_{k+1}) \Big),
\end{equation}
such that $\tilde{V}^{\pi}(x_k, z_k) = \Eb_{u_k}\big[ \tilde{Q}^{\pi}(x_k, z_k, u_k) \big]$.
We can now derive a policy gradient theorem for the inner problem.
\begin{mdframed}[style=GreyFrame]
\begin{restatable}[Policy Gradient Theorem]{thm}{}\label{thm:policy_gradient}
The gradient of the policy value function $\tilde{V}^{\pi_\theta}$ \eqref{eq:stoch_V_efocp:def} for the inner subproblem satisfies
\begin{equation} \label{eq:policy_gradient}
\begin{split}
    &\mathrel{\phantom{=}}\nabla_\theta \tilde{V}^{\pi_\theta}(x_0, z_0) \\
    &\propto \E_{(x, z, u)_{1:k} \sim \pi_\theta}\Big[ \tilde{Q}^{\pi_\theta}(x_k, z_k, u_k) \nabla_\theta \ln \pi_\theta(u_k | x_k, z_k) \Big| \xi=1 \Big], \raisetag{2.1\baselineskip}
\end{split}
\end{equation}
where the binary random variable $\xi_k$ is defined to be equal to $1$ when $h(x_t) \leq \tilde{V}^{\pi_\theta}(x_{t+1}, z_{t+1})$ is true for all $t = 0, \dots, k$.
\end{restatable}
\end{mdframed}
\begin{IEEEproof}
The proof follows from the proof of the normal policy gradient theorem \cite{sutton2018reinforcement}, differing only in the expression for $\nabla_\theta \tilde{Q}^{\pi_\theta}$. In the normal setting \cite{sutton2018reinforcement}, \eqref{eq:stoch_V_sum:dp} gives
\begin{equation}
    \nabla_{\theta} Q^{\pi_\theta}(x_k, u_k) = \nabla_{\theta} \Big( l(x_k) + V^{\pi_\theta}( x_{k+1} ) \Big)
    = \nabla_{\theta} V^{\pi_\theta}( x_{k+1} ).
\end{equation}
In the case of EF-OCP with \eqref{eq:stoch_V_efocp:dp},
\begin{align}
    \nabla_{\theta} \tilde{Q}^{\pi_\theta}(x_k, u_k)
    &= \nabla_{\theta} \max\Big\{ h(x_k), \tilde{V}^{\pi_\theta}( x_{k+1}, z_{k+1} )  \Big\}, \\
    &= \ind_{h(x_k) \leq \tilde{V}^{\pi_\theta}(x_{k+1}, z_{k+1})} \nabla_{\theta} \tilde{V}^{\pi_\theta}( x_{k+1}, z_{k+1} ).
\end{align}
Following the rest of the normal proof \cite{sutton2018reinforcement} then yields \eqref{eq:policy_gradient}
\end{IEEEproof}
\smallskip

From \Cref{thm:policy_gradient}, we can construct a basic on-policy DeepRL algorithm to solve the inner problem \eqref{eq:opt:sa_inner} over all states $x$ parametrized by a range of $z$, yielding a learned (stochastic) controller $\pi_\theta(x, z)$. However, we can do better by performing variance reduction via subtracting the baseline $\tilde{V}^{\pi_\theta}(x, z)$ to get the advantage $\tilde{A}^{\pi_\theta}(x, z, u) \coloneqq \tilde{Q}^{\pi_\theta}(x, z, u) - \tilde{V}(x, z)$, as $\E_{\pi_\theta}[\nabla_\theta \ln \pi_\theta] = 0$. Following PPO, we also apply the GAE estimate \cite{schulman2015high}, perform clipped importance sampling and add in an entropy bonus to arrive at an algorithm that is very similar to PPO \cite{schulman2017proximal} but with $\tilde{Q}^\pi$ and $\tilde{V}^\pi$ defined as above.

\noindent\textbf{Stochastic policy considerations: }
While we have performed the above developments using a stochastic policy and derived a stochastic policy gradient theorem, our desired solution to the inner optimization problem is a \textit{deterministic} controller. Hence, we only take the mode of the learned policy, and treat the stochasticity purely as a means of performing exploration, discouraging premature convergence to local minima and smoothing the optimization landscape \cite{mnih2016asynchronous,haarnoja2018soft,ahmed2019understanding}.
We also fine-tune the learned value function at the end of the DeepRL training by freezing the obtained deterministic policy and performing policy evaluation.

It can be appealing to consider using a deterministic policy and apply a \textit{deterministic} policy gradient theorem \cite{silver2014deterministic}. However, the proof of this theorem requires the transition distribution $p(x_{t+1} | x_t, u_t)$ to be continuous \cite{silver2014deterministic}. This does not hold in our problem since the transition distribution is degenerate (and hence discontinuous) due to the use of deterministic dynamics, i.e.,
\begin{equation}
    p(x_{t+1} | x_t, u_t) = \delta\big( x_{t+1} - f(x_t, u_t) \big),
\end{equation}
preventing the use of deterministic policy gradient in this case.

After obtaining $\tilde{V}^\pi$ and $\pi$ for the deterministic policy, we now turn to the outer problem \eqref{eq:opt:epi1}.
Since the outer optimization problem is only $1$ dimensional, we can solve for $z^*$ easily via classical scalar optimization methods such as the bisection method which run quickly. Instead of running bisection online, however, we choose to learn the optimal $z^* : \XSet \to \Rb$ offline by using the result of bisection as the label for a regression problem. This gives us $z^*(x)$, which in turn provides the optimal policy for the original constrained optimal control problem.

The proposed EFPPO algorithm is summarized in \Cref{alg:efppo:inner,alg:efppo:outer} and illustrated in \Cref{diag:efppo}.
EFPPO solves the two-stage optimization problem of EF-COCP \eqref{eq:opt:efcocp} sequentially. The inner problem \Cref{eq:opt:sa_inner} is solved via policy gradient using \Cref{thm:policy_gradient} for a range of $z$ values and uses the improvements in PPO such as GAE estimates, clipped importance sampling, and an entropy bonus. Next, we extract the mode of the stochastic $\pi$ and fine-tune $\tilde{V}^{\pi}$. Then, we fix the value function $\tilde{V}^\pi$ and the deterministic policy $\pi(x, z)$ and learn $z^*(x)$ by randomly sampling states in the state space and minimizing the residual to the analytical solution of \eqref{eq:opt:epi1} found by applying the bisection method. Since this is a $1D$ optimization problem, the bisection method converges to almost machine precision within tens of iterations. The final policy is then obtained as $\pi(x, z^*(x))$.
When solving the inner problem, we randomly sample states from the state space and random sample $z$ within the range $[0, z_{\max}]$, where $z_{\max}$ is an upper-bound estimate of the total cost $\sum_{k=0}^\infty l(x_k)$. To prevent using $z_{\max} = \infty$ since this term may be unbounded, we introduce a small discount factor only in the total costs. More details on the effect of discounting are available in the \CrefApp{app:sec:discounts}.

\begin{algorithm}[t]
    \small
    \caption{EFPPO Inner Problem}
    \label{alg:efppo:inner}
    \begin{algorithmic}
        \State{\textbf{input: }Estimate of maximum cost upper bound $z_{\max}$}
        \State
        \Repeat
            \If{reset environments}
                \State Sample random $x_0 \in \XSet$ and $z_0 \in [0, z_{\max}]$
            \EndIf
            \State Run policy $\pi$ in environments for $T$ timesteps
            \State Compute advantage estimates $A^\pi$ with GAE
            \State Update policy $\pi$ and baseline $\tilde{V}^\pi$ with PPO clipping and entropy
        \Until{converged}
        \State Fine-tune $\tilde{V}^{\pi}$ using the mode of $\pi$ via policy evaluation.
    \end{algorithmic}
\end{algorithm}
\begin{algorithm}[t]
    \small
    \caption{EFPPO Outer Problem}
    \label{alg:efppo:outer}
    \begin{algorithmic}
        \State{\textbf{input: }Estimate of maximum cost upper bound $z_{\max}$}
        \State
        \State Sample dataset of $x$ randomly from the state space
        \State Bisect $\tilde{V}^\pi(x, \cdot)$ over $[0, z_{\max}]$ to obtain labels $z^*$ for each $x$
        \State
        \Repeat
            \State Train network $\tilde{z}^*(x)$ to predict $z^*$ given $x$ via regression
        \Until{converged}
    \end{algorithmic}
\end{algorithm}

\section{Experiments}
To evaluate the performance of the proposed EFPPO algorithm, we compare EFPPO against related algorithms on simulated stabilize-avoid problems with increasing complexity.
The last problem involves a non-differentiable, nonconvex, non-control-affine system and demonstrates the ability of our approach to both maintain safety and successfully stabilize the system within the goal region even in nontrivial high-dimensional environments. We compare EFPPO against the following baselines methods.
\begin{itemize}
\item PPO \cite{schulman2017proximal}, a popular on-policy \textit{unconstrained} DeepRL method. Despite its unconstrained nature, it is common to apply ``soft constraints'' in the form of penalties incurred when an undesirable state is reached \cite{brockman2016openai,wang2021pretrained}. Unlike constrained methods such as EFPPO, the scale of the penalties is a hyperparameter that must be tuned to trade-off between constraint violation and training stability. We denote by PPO($\lambda$) the \textit{family} of methods where $\lambda$ denotes the penalty scale of the modified cost function $\tilde{l}$, i.e.,
    \begin{equation}
        \tilde{l}(x) = l(x) + \lambda [h(x)]^{+},
    \end{equation}
    where PPO($0$) solves the unconstrained problem.
\item CPPO \cite{stooke2020responsive}, a representative algorithm among the family of constrained DeepRL methods which use Lagrangian duality, and is an improvement on PPO-Lagrangian from \cite{achiam2017constrained}. As noted in \Cref{subsec:lagr_comparison}, the constrained MDP formulation allows constraint violations up to a cost threshold. To learn policies that strictly satisfy the constraints, we set the cost threshold to $0$. We perform a manual hyperparameter search to select the PID parameters.
\item PPO-SIS \cite{ma2022joint}, a constrained RL method that applies the safety constraint at each state instead of in expectation as in CMDP. Moreover, a safety certificate is learned jointly to improve the safety of the learned policy.
\item CLBF\footnote{While the original work considers CLBFs that are robust to parametric uncertainties, the problems we consider here do not have parameters.} \cite{dawson2022safe}, which learns a Lyapunov function using a neural network via losses which penalize violations of the Lyapunov conditions. Safety can be guaranteed by enforcing that a sublevel set of the learned neural Lyapunov function lies outside the avoid set. Given a learned CLBF, we consider two methods for synthesizing the controller: (1) CLBF(QP) solves the CLBF-QP using the CLBF as a constraint; (2) CLBF(Opt) applies the (bang-bang) control that minimizes $\dot{V}$.
\end{itemize}

As noted previously, while the reach-avoid problem is closely related to the stabilize-avoid problem considered in this work, the policies obtained from reach-avoid do not induce stability when the goal set is not an equilibrium point (see \Cref{subsec:reachavoid_relationship}).

For a fair comparison, a feedforward neural network with $\tanh$ activations is used for both the policy (if used) and value functions. Additional networks are also defined using the same architecture above but with modified final activation function (e.g., softplus) when used in the original implementation.

To compare each method, we use the safety rate, cost and stabilize rate metrics computed by rolling out the learned policy on randomly sampled initial states. When the true control-invariant set that can guarantee safety is analytically known, the sampled states are sampled from the control-invariant set. However, for systems with complex dynamics where such a set cannot be found analytically, we sample from a crude box estimate. The safety rate and stabilize rate are defined as the fraction of trajectories that satisfy the safety constraints and can reach and stay within the goal set for the last $50$ timesteps respectively.
The top method for each metric is shown in \textbf{bold}. For the cost, we only highlight the best performing method among the methods that handle constraints (i.e., unconstrained PPO is not taken into account).

More details regarding the specific dynamics and constraints for each task are provided in the \CrefApp{app:sec:exp_details} for brevity.

\subsection{1D Double-Integrator}
\begin{figure*}[!t]
\centering
\includegraphics[width=\linewidth,draft=false]{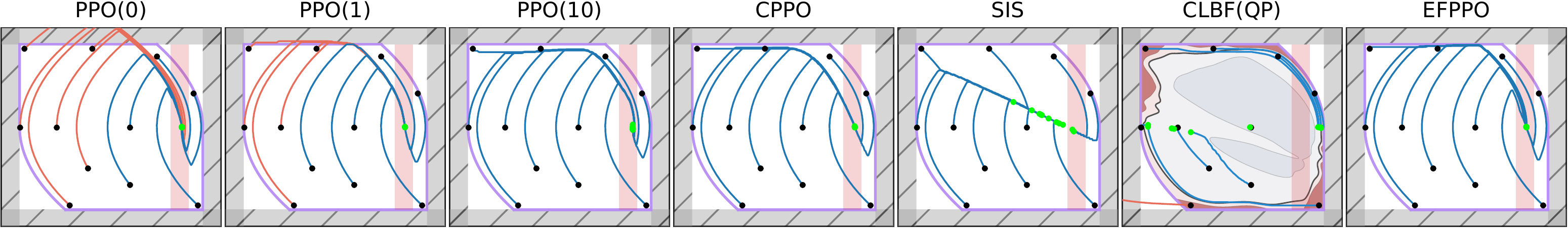}
\captionof{figure}{Trajectory rollouts (\mycirc[black]{} $\textcolor{tabblue}{\boldsymbol{\to}}$ \mycirc[green]{}) on the double-integrator system with box constraints on both position and velocity (shown in grey) for the avoid set. Unsafe trajectories are shown in \textcolor{tabred}{red}.
For CLBF(QP), a contour plot of the learned CLBF is shown with \textcolor{red}{red} regions denoting higher values and the safe level set shown as the \textcolor{tabblue}{blue} region.
Due to control constraints $\abs{u} \leq 1$, the control-invariant region \includegraphics[height=1.5ex,draft=false]{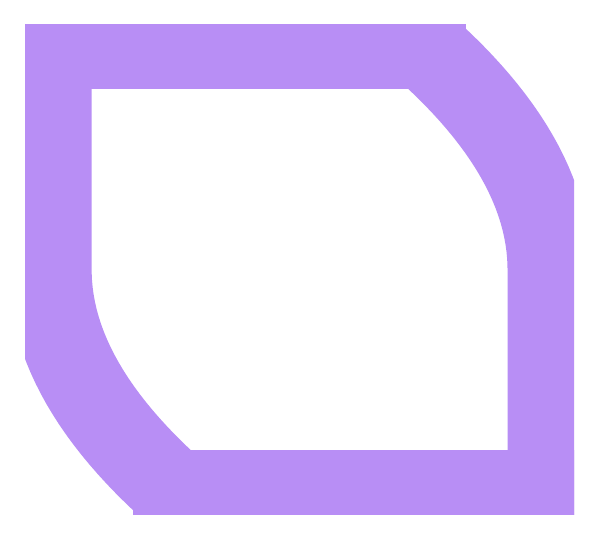} is smaller than the complement of the avoid set.}
\label{fig:twodint} \end{figure*}

We first consider a simple double-integrator dynamics in 1D, where the optimal policy and corresponding optimal control-invariant region for the problem can be computed. Given states $[p, v] \in \Rb^2$, controls $u = a \in \Rb$ the task here is to stabilize to the region $p_{\text{goal}} \coloneqq [0.65, 0.85]$ while satisfying box constraints on both states and controls
\begin{equation}
    \abs{p} \leq 1, \quad \abs{v} \leq 1, \quad \abs{u} \leq 1.
\end{equation}
While all constrained methods are able to maintain safety for all states within the control-invariant region, only EFPPO is able to also stabilize all trajectories to the goal set $\GoalSet$ with a cost similar to the unconstrained solution from PPO(0) (see \Cref{tab:twodint}).
PPO-SIS has regions of the state space outside $\GoalSet$ which are equilibrium points (see \Cref{fig:twodint}), while CPPO has a much larger cost compared to the unconstrained solution.

The unconstrained PPO($\lambda$) methods require the penalty weight $\lambda$ to be large enough for a safe policy, but this trade-off comes at the cost of the policy's performance and requires careful tuning of the penalty weight $\lambda$. In contrast, EFPPO is able to synthesizes a performant safe controller without requiring any cost function tuning.

\begin{table}[t]
{
\def \cppoLabel {CPPO \cite{stooke2020responsive}}
\def \sisLabel {PPO-SIS \cite{ma2022joint}}
\def \clbfQPLabel {CLBF(QP) \cite{dawson2022safe}}
\def \clbfOptLabel {CLBF(Opt) \cite{dawson2022safe}}
\centering
\begin{booktabs}{
  colspec = {lccc},
}
\toprule 
  & {Safety\\ Rate \uarr} & Cost \darr  & {Stabilize\\ Rate \uarr} \\ \midrule
PPO($0$)   & $0.55$                   & $1.251$      & $\bm{1.0}\hphantom{00}$ \\
PPO($1$)   & $0.61$                   & $1.283$      & $\bm{1.0}\hphantom{00}$ \\
PPO($10$)  & $\bm{1.0}\hphantom{0}$   & $1.299$      & $\bm{1.0}\hphantom{00}$ \\
\addlinespace
\cppoLabel & $\bm{1.0}\hphantom{0}$   & $1.314$      & $\bm{1.0}\hphantom{00}$ \\
\sisLabel  & $\bm{1.0}\hphantom{0}$   & $2.227$      & $0.182$ \\
\clbfQPLabel  & $0.96$   &  $3.042$   & $0.180$ \\
\clbfOptLabel & $0.99$   & $3.180$    & $0.035$ \\
\addlinespace
\SetRow{MaterialBlue50}
EFPPO (Ours)  & $\bm{1.0}\hphantom{0}$   & $\bm{1.285}$ & $\bm{1.0}\hphantom{00}$ \\
\bottomrule
\end{booktabs}
\caption{
    Comparison of controller performance on the double integrator example. 
    Metrics are computed from states randomly sampled from the control-invariant set.
}
\label{tab:twodint}
} \end{table}

\begin{figure*}[!t]
\centering
\includegraphics[width=\linewidth,draft=false]{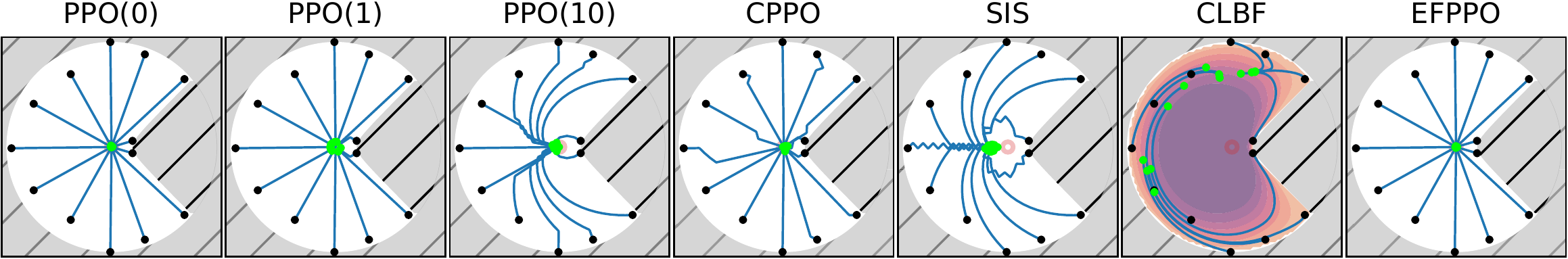}
\caption{Trajectory rollouts (\mycirc[black]{} $\textcolor{tabblue}{\boldsymbol{\to}}$ \mycirc[green]{}) on the single-integrator system in 2D with sector constraints.
For CLBF, a contour plot of the learned CLBF is shown.
}
\label{fig:sectorcircle} \end{figure*}

{
\def \cppoLabel {CPPO \cite{stooke2020responsive}}
\def \sisLabel {PPO-SIS \cite{ma2022joint}}
\def \clbfQPLabel {CLBF(QP) \cite{dawson2022safe}}
\def \clbfOptLabel {CLBF(Opt) \cite{dawson2022safe}}
\begin{table}[t]
\centering
\begin{booktabs}{
  colspec = {lccc},
}
\toprule 
  & {Safety\\ Rate \uarr} & Cost \darr  & {Stabilize\\ Rate \uarr} \\ \midrule
PPO($0$)      & $\bm{1.0}$   & $0.363$      & $\bm{1.0}$ \\
PPO($1$)      & $\bm{1.0}$   & $0.364$      & $\bm{1.0}$ \\
PPO($10$)     & $\bm{1.0}$   & $0.459$      & $\bm{1.0}$ \\
\addlinespace
\cppoLabel    & $\bm{1.0}$   & $0.365$      & $\bm{1.0}$ \\
\sisLabel     & $\bm{1.0}$   & $2.626$      & $0.0$ \\
\clbfQPLabel  & $\bm{1.0}$   & $12.613$     & $0.0$ \\
\clbfOptLabel & $\bm{1.0}$   & $1.308$      & $0.0$ \\
\addlinespace
\SetRow{MaterialBlue50}
EFPPO (Ours)  & $\bm{1.0}$   & $\bm{0.363}$ & $\bm{1.0}$ \\
\bottomrule
\end{booktabs}
\caption{
    Comparison of controller performance on the 2D single integrator with sector obstacles. Metrics are computed from states randomly sampled from the control-invariant set.
}
\label{tab:sectorcircle}
\end{table}
}

\subsection{2D Single-Integrator with Sector Obstacle}
Next, we consider a single-integrator in 2D where the avoid set is defined as a sector of the circle and the goal set defined in the center. The state is defined as the positions $[p_x, p_y] \in \Rb^2$ with controls denoting the velocities $[v_n, v_t] \in [-1, 1]^2 \subset \Rb^2$, where $v_n$ and $v_t$ denote the normal and tangent components of the velocity vector to the center of the circle.

\Cref{tab:sectorcircle} summarizes the results in this task.
Due to the control parametrization and the shape of the avoid set, the optimal control at all states is the constant vector $[1, 0]$. Of the constrained methods, only EFPPO is able to learn this, with other methods learning suboptimal versions of the optimal policy.
Note that the rollouts for CLBF \cite{dawson2022safe} actually move away from the center when starting near the sector obstacle on the right (see \Cref{fig:sectorcircle}). This is because the CLBF learns a distorted metric near the obstacles (as seen from the value function) to ensure the level-set is contained within the control-invariant set.
Moreover, we see that SIS also fails to reliably stabilize on this simple example. The safety index used in SIS \cite{ma2022joint} is taken from \cite{zhao2021model} and consists of only three parameters. We suspect that this is insufficient for tasks where the avoid set $\AvoidSet$ is relatively complex.
Finally, we see that PPO($10$) learns a particularly poor performing controller due to the large penalty weights. In particular, the difference in scales between the large costs in the unsafe regions and the small costs near the goal set poses a challenge for learning the value function accurately, resulting in the learned policy putting more weight on constraint satisfaction.

\begin{figure*}[t]
{
\def\myfrac{0.195\linewidth}
\centering
\begin{subfigure}[b]{\myfrac}
    \centering
    \includegraphics[width=\linewidth]{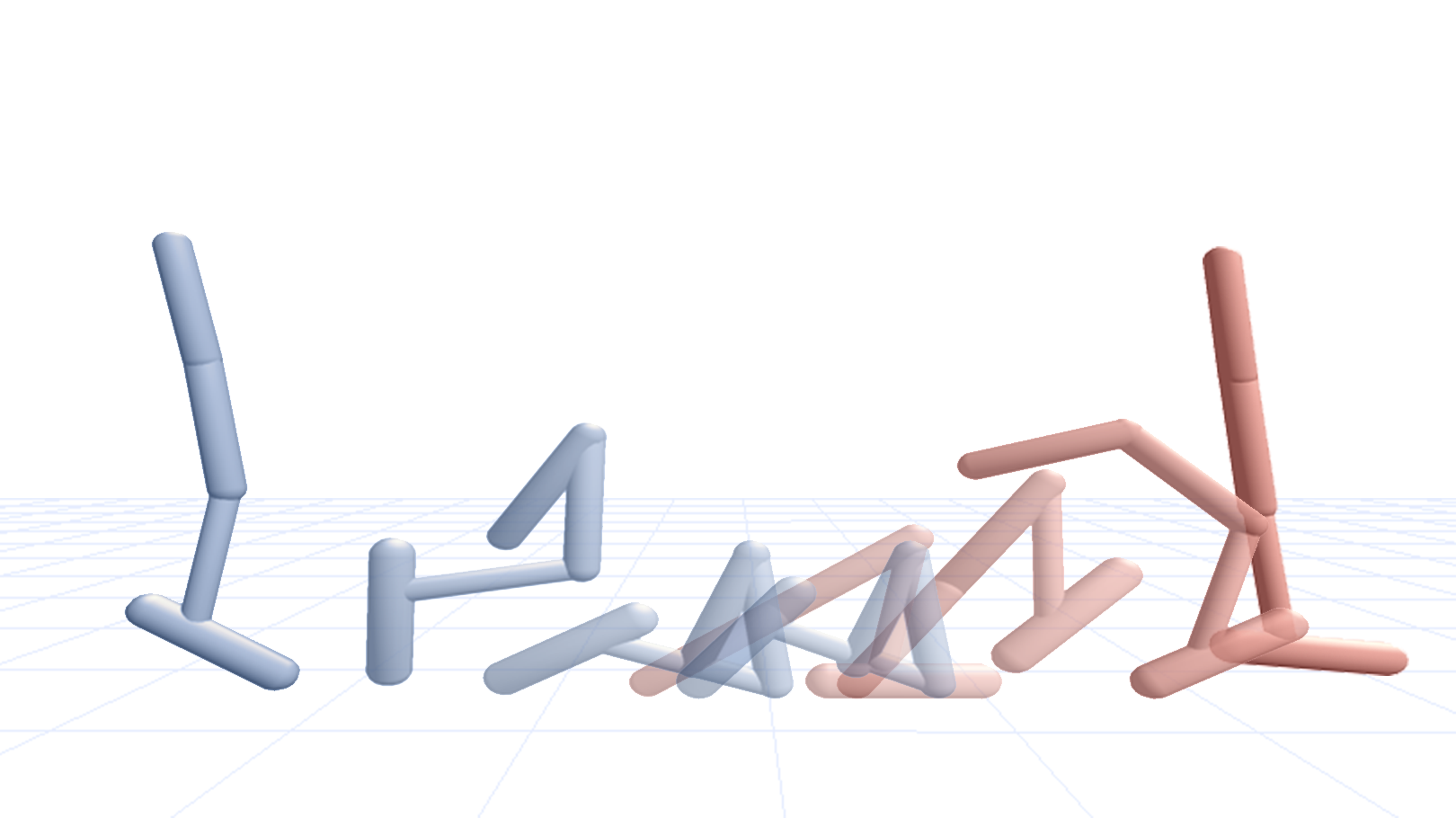}\caption{PPO($0$)}
\end{subfigure}
\hfill
\begin{subfigure}[b]{\myfrac}
    \centering
    \includegraphics[width=\linewidth]{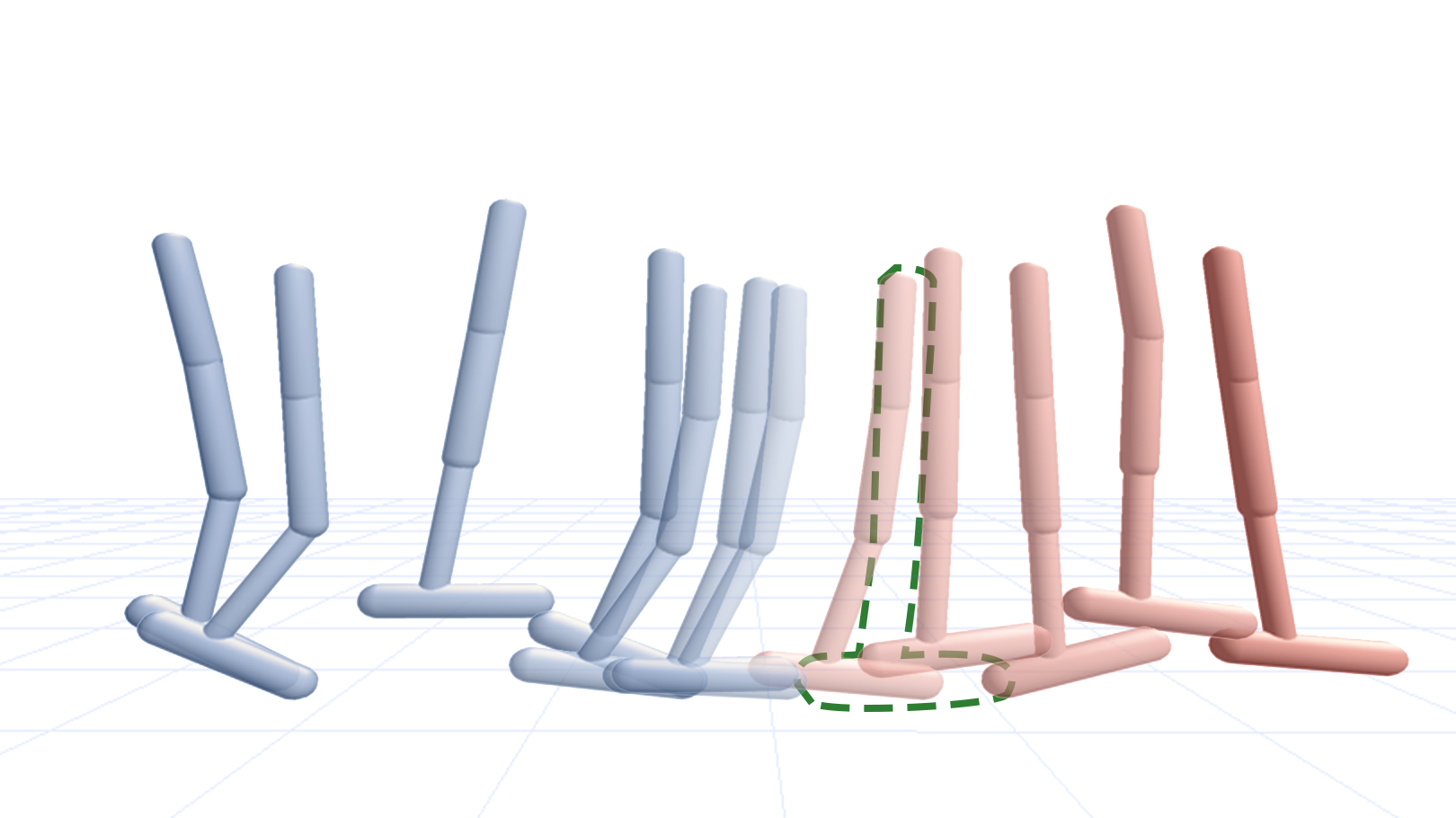}\caption{PPO($1$)}
\end{subfigure}
\hfill
\begin{subfigure}[b]{\myfrac}
    \centering
    \includegraphics[width=\linewidth]{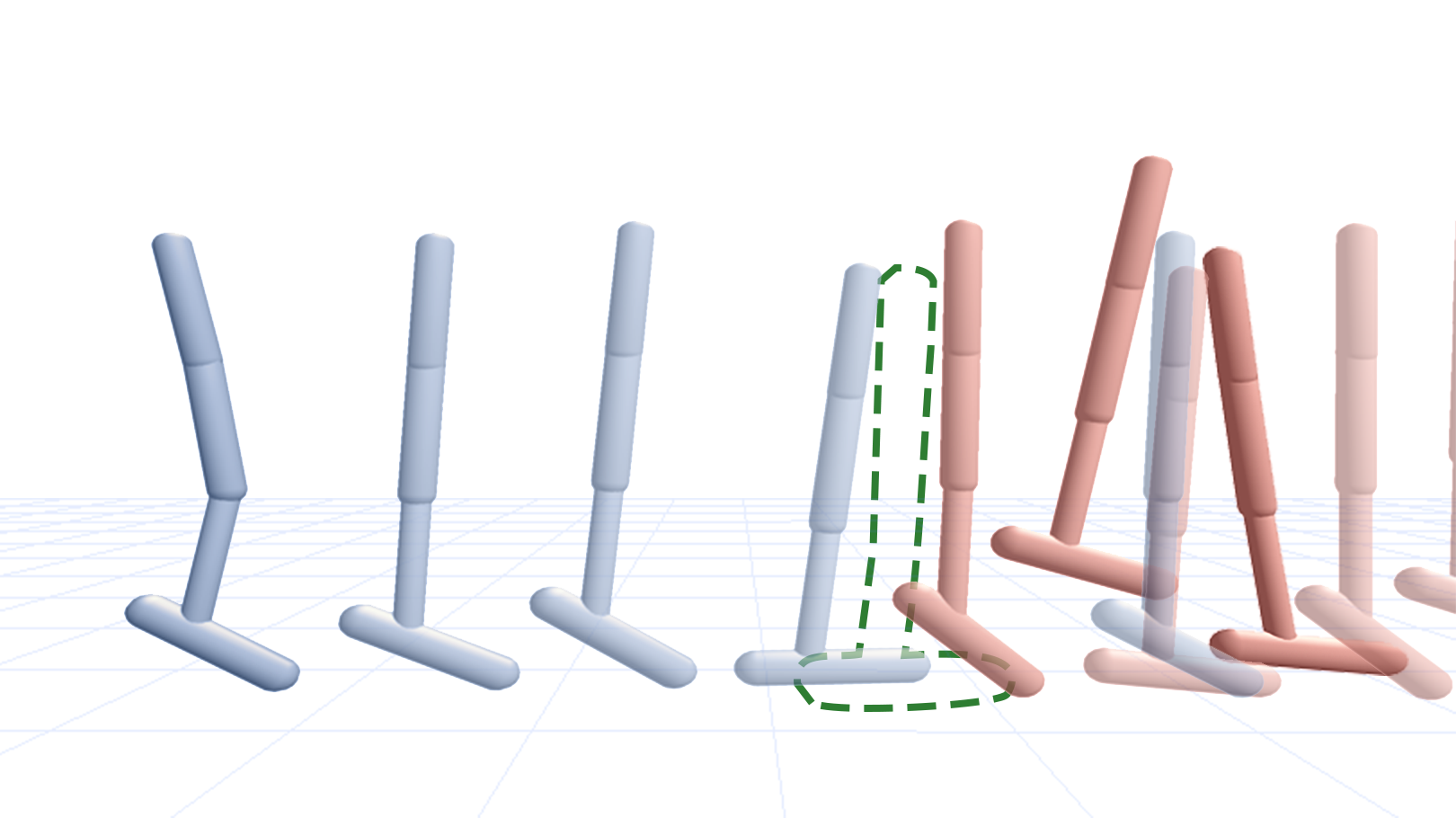}\caption{PPO($10$)}
\end{subfigure}
\hfill
\begin{subfigure}[b]{\myfrac}
    \centering
    \includegraphics[width=\linewidth]{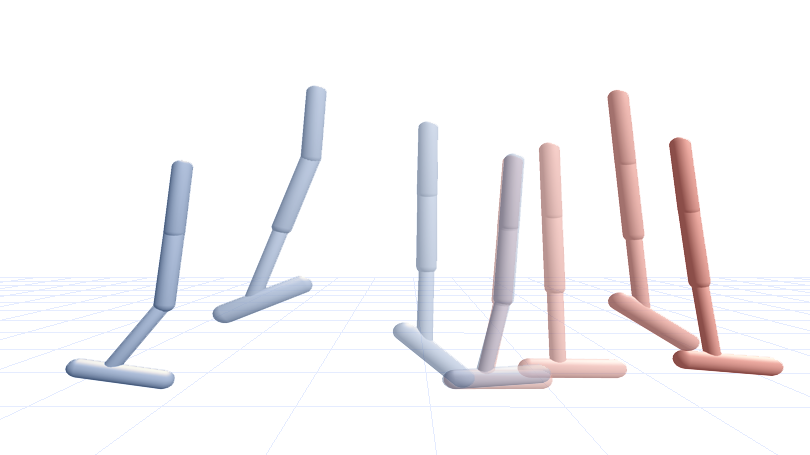}\caption{EFPPO}
\end{subfigure}
\hfill
\begin{subfigure}[b]{\myfrac}
    \centering
    \includegraphics[width=\linewidth]{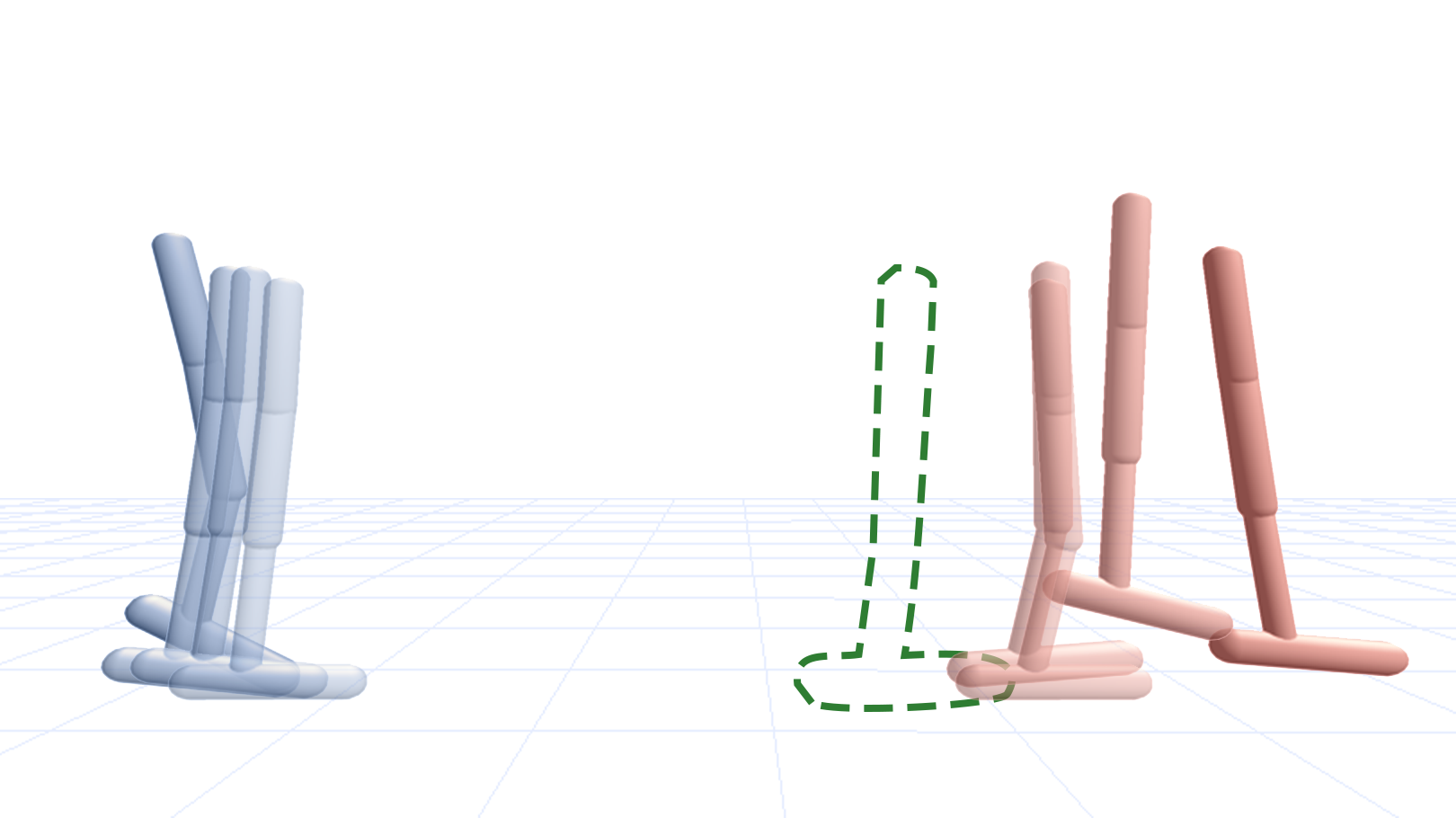}
    \caption{CPPO} \label{fig:hopper:cppo}
\end{subfigure}\caption{The hopper task asks for stabilization of the hopper's torso to the set $p_x \in [2.8, 3.0]$ (dashed green lines, drawn when the hopper does not reach the goal set). Two rollouts are shown starting from $p_x=1.0$ (blue) and $p_x=4.0$ (red) respectively, with time progressing from opaque to transparent. Only EFPPO safely stabilizes the system to the goal set in these two rollouts.
}
\label{fig:hopper}
} \end{figure*}
\subsection{Hopper Stabilization}
The preceding benchmarks are low-dimensional toy-examples that provide intuition on how the different methods behave for simple linear systems. Our next example demonstrates the ability of EFPPO to stabilize to goal sets while maintaining safety for more complex dynamics. We consider the hopper in the Brax simulator \cite{brax2021github}, a nonlinear non-differentiable system with a $12$ dimensional state space and a $3$ dimensional control space. Note that the original dynamics only consider a $11$ dimensional state space as they discard the x-coordinate, which we keep.
The goal set $\GoalSet$ in this task is defined as
\begin{equation}
    \GoalSet \coloneqq \Set{ \vx | p_x \in [2.8, 3.0] },
\end{equation}
while the constraints limit the height and rotation of the hopper's torso
\begin{equation}
    p_z \geq 0.7, \qquad \abs{\theta} \leq 0.2.
\end{equation}
Unlike the normal setup for Hopper, stabilization to $\GoalSet$ requires keeping track of the x-position. Consequently, the optimal policy is no longer a limit-cycle and even requires the hopper to move backwards for some initial states. We report our results in \Cref{tab:hopper}. On this problem, we see that both CPPO and EFPPO achieve high safety rates, while PPO-SIS fails to stabilize. We see the same trend of PPO($\lambda$) presenting a trade-off between stability and safety, except for PPO($10$) which is more unsafe than PPO($1$). We suspect this is due to larger costs destabilizing training.

{
\def \cppoLabel {CPPO \cite{stooke2020responsive}}
\def \sisLabel {PPO-SIS \cite{ma2022joint}}
\def \hpz {\hphantom{0}}
\begin{table}[t!]
\centering
\begin{booktabs}{
  colspec = {lcccc},
  cell{1}{2} = {r=2}{m}, cell{1}{3} = {c=2}{c}, cell{1}{5} = {r=2}{m}, }
\toprule 
  & {Safety\\ Rate \uarr} & Cost \darr &   & {Stabilize\\ Rate \uarr} \\ \cmidrule[l]{3-4}
  &  & Safe & All  &  \\ \midrule
PPO($0$)      & $0.000$      &                  & $\hpz1.360$      & $0.893$ \\
PPO($1$)      & $0.676$      & $\hpz2.986$      & $\hpz3.806$      & $0.661$ \\
PPO($10$)     & $0.037$      & $15.193$         & $15.503$         & $0.087$ \\
\addlinespace
\cppoLabel    & $0.724$      & $\hpz8.353$      & $\hpz9.458$      & $0.087$ \\
\sisLabel     & $0.000$      &                  & $\hpz9.703$      & $0.084$ \\
\addlinespace
\SetRow{MaterialBlue50}
EFPPO (Ours)  & $\bm{0.833}$ & $\hpz\bm{1.568}$ & $\hpz\bm{3.695}$ & $\bm{0.843}$ \\
\bottomrule
\end{booktabs}
\caption{
    Comparison of controller performance on the Hopper system on a set of random initial states that may lie outside the (unknown) optimal control-invariant set.
}
\label{tab:hopper}
\end{table}
} 
\subsection{F16 Ground Collision Avoidance In a Low Altitude Flight Corridor}
\begin{figure*}[t!]
    \centering
\begin{subfigure}[b]{0.49\linewidth}
        \centering
        \includegraphics[width=\linewidth,trim={16cm 2.0cm 5cm 3.5cm},clip]{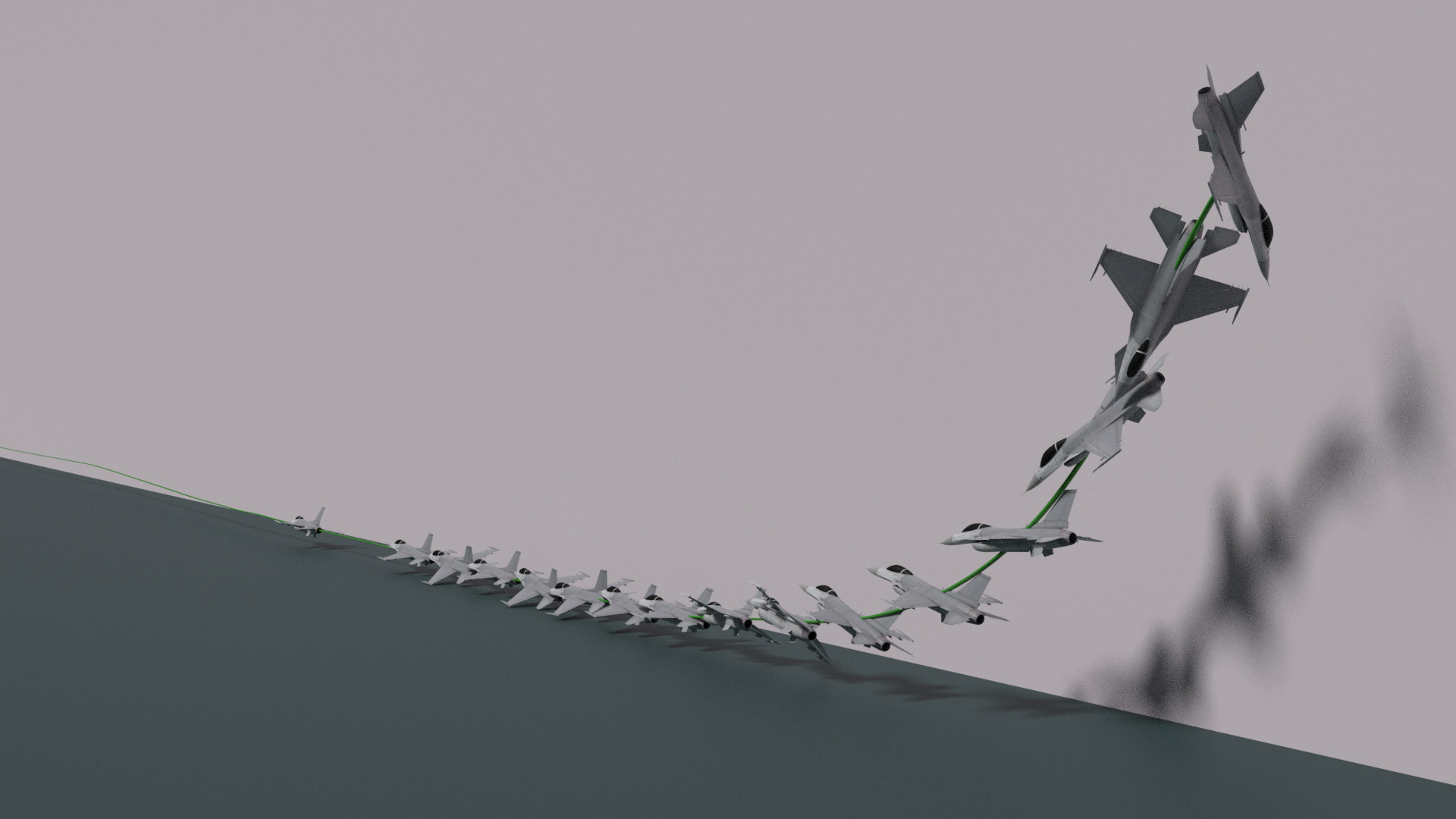}
        \caption{EF-PPO}
    \end{subfigure}\hfill \begin{subfigure}[b]{0.49\linewidth}
        \centering
        \includegraphics[width=\linewidth,trim={16cm 2.0cm 5cm 3.5cm},clip]{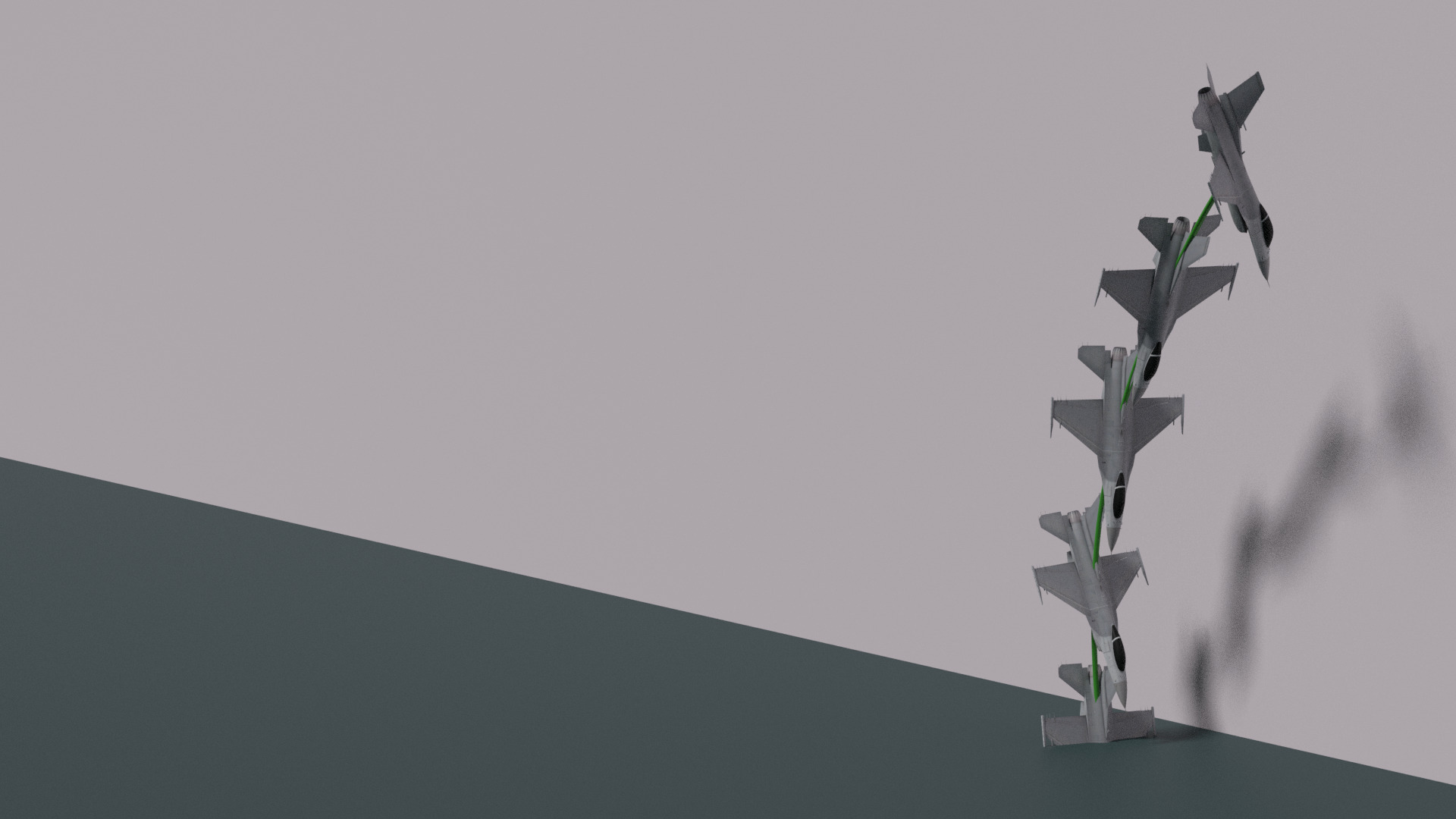}
        \caption{CPPO}
    \end{subfigure}
    \caption{Ground collision avoidance and stabilization to a low altitude flight corridor for the F16 fighter jet system.}
    \label{img:f16}
\end{figure*}

Finally, we showcase the scalability of our method on a ground collision avoidance example involving the F16 fighter jet \cite{heidlauf2018verification}. This system is non control-affine, non-smooth and involve lookup tables, making this a challenging system to solve stabilize-avoid on in addition to the high $17$-dimensional state space and $4$-dimensional control space. The task here is to stabilize the F16 to a target altitude defined by the set $[50, 150] \unit{\ft}$ under box control constraints while staying within a flight corridor heading North (box constraints on the East-Up plane). For positions $(p_E, p_N, p_U)$, this corresponds to
\begin{equation}
    \qty{-200}{\ft} \leq p_E \leq \qty{200}{\ft}, \quad
    \qty{0}{\ft} \leq p_U \leq \qty{1000}{\ft}.
\end{equation}
The results are summarized in \Cref{tab:f16} and \Cref{img:f16}.
Following the trend from before, we see that CPPO is able to achieve high safety rates at the expense of being unable to stabilize to the goal set. In contrast, EFPPO using the optimal $z^*$ achieves similar safety rates as CPPO but has a $70$-fold increase in the stabilize rate. Moreover, note that EFPPO($0$) and EFPPO($1.8$) denote using a constant value of $z$ for the policies from the inner EF-COCP problem \eqref{eq:opt:sa_inner}. As expected, using a larger value of $z$ results in improved constraint satisfaction at the cost of a lower stabilize rate. By using the optimal value of $z^*$, EFPPO($z^*$) is able to achieve the high stabilize rate of EFPPO($0$) while maintaining the high safety rates of EFPPO($1.8$). However, note that the safety rate of EFPPO($z^*$) is actually slightly lower than that of EFPPO($1.8$). This is because $z^*$ is approximated via a neural network which may have small approximation errors.
We also see the same trends for PPO($\lambda$) from Hopper carry over to this task.

 {
\def \cppoLabel {CPPO \cite{stooke2020responsive}}
\begin{table}[t]
\centering
\begin{booktabs}{
  colspec = {lcccc},
  cell{1}{2} = {r=2}{m}, cell{1}{3} = {c=2}{c}, cell{1}{5} = {r=2}{m}, }
\toprule 
  & {Safety\\ Rate \uarr} & Cost \darr &   & {Stabilize\\ Rate \uarr} \\ \cmidrule[l]{3-4}
  &  & Safe & All  &  \\ \midrule
PPO($0$)                & $0.000$      &              & $0.843$      & $0.917$ \\
PPO($1$)                & $0.657$      &  $0.639$     & $1.025$      & $0.742$ \\
PPO($10$)               & $0.000$      &              & $4.032$      & $0.092$ \\
\addlinespace
\cppoLabel            & $0.827$      & $4.874$      & $4.822$      & $0.012$ \\
\addlinespace
\SetRow{MaterialBlue50}
EFPPO($0$) \hspace{0.8em}(Ours)     & $0.792$ & $0.735$      & $\bm{1.102}$ & $0.837$ \\
\SetRow{MaterialBlue50}
EFPPO($1.8$) (Ours)   & $\bm{0.856}$      & $2.806$      & $2.945$      & $0.022$ \\
\SetRow{MaterialBlue50}
EFPPO($z^*$) \hspace{0.3em}(Ours)   & $0.823$      & $\bm{0.724}$ & $1.139$      & $\bm{0.843}$ \\
\bottomrule
\end{booktabs}
\caption{
    Comparison of controller performance on the F16 system on a set of random initial states that may lie outside the (unknown) optimal control-invariant set.
}
\label{tab:f16}
\end{table}
}  \section{Discussion} \label{sec:discussion}
Many Lagrangian duality-based methods suffer from focusing too much on satisfying the safety constraints and consequently suffer in terms of their stability. As noted in \Cref{subsec:lagr_comparison}, the non-negative gradients of the Lagrange multipliers $\lambda$ means that they will continue to increase as long as constraints are not being satisfied. However, a large $\lambda$ causes the optimization problem to become badly conditioned, hindering the performance of these algorithms. Moreover, in systems with complex dynamics where the control-invariant set is not known, it may be impossible for constraints satisfaction to occur in all states. In this case, $\lambda$ will grow unbounded until either the training algorithm becomes unstable or it hits a user-defined maximum limit.
This is a failure that the PID mechanism in CPPO does not fix, as $\lambda$ does not oscillate. We found all three coefficients of the PID played a similar role in adjusting the rate at which $\lambda$ (monotonically) increases and eventually destabilizes training.
Although the CMDP problem formulation can be adapted to solve stabilize-avoid problems, these methods are a poor fit.

Additionally, we observe that the CLBF performs poorly even on both toy examples.
We believe there are two reasons for this.
\begin{enumerate}
    \item Learning of the CLBF assumes that the control-invariant set is known (or can be sampled from) a-priori. However, in the experiments considered in this work, we do not assume knowledge of these sets. While we do give CLBF a small region around the equilibrium subset of the goal region, information that was not available to other algorithms, this was not enough for CLBFs to perform well.
    \item The learning problem formulated in \cite{dawson2022safe} is \textit{underparametrized}. More specifically, given any CLBF $V$ which satisfies the CLBF conditions, $\alpha V$ will induce the same set of feasible constraints and will perform identically to $V$. However, if $V$ does not satisfy the CLBF conditions, the violation error can be reduced arbitrarily by taking $\alpha \to 0$. This yields a value function where regions close to the goal set have near-zero gradients which may violate the CLBF conditions and compromise the controller's performance, which we have observed empirically in our experiments.
\end{enumerate}

In contrast, the proposed EFPPO algorithm can perform well even when large areas of the state space are not control-invariant as in our experiments. This is due to two reasons: (a) $z$ does not affect the magnitude of the gradients directly, unlike $\lambda$ which scales the gradients of the constraint (and also the cost, if the reweighing scheme from \cite{stooke2020responsive} is used). Consequently, while $z \to \infty$ will make no difference to EFPPO (if the $z$ feature variable is normalized correctly), taking $\lambda \to \infty$ will cause the Lagrangian to diverge and cause training instabilities. (b) Since $z$ can be interpreted as a ``cost budget'' and it is easier to estimate an upper-bound on costs, we can afford to solve the two-stage optimization problem sequentially. In contrast, it is much harder to bound the optimal $\lambda$ since the units of ``cost to constraint ratio'' is more difficult to reason about.

\section{Conclusion}
We present a new method for synthesizing nonlinear feedback controllers for performing stabilization while maintaining safety under control constraints. By formulating the stabilize-avoid problem as an infinite-horizon epigraph-form constrained optimal control problem and applying deep reinforcement learning, our approach is able to sidestep numerical challenges that other methods face and achieve vastly larger regions of attraction while still maintaining safety for high dimensional complex systems. 

\vspace{0.18\baselineskip}

\noindent\textbf{Limitations and future work: }
The EFPPO algorithm currently splits the task of learning $V, \pi$ and $z^*$ into separate stages of optimization, and relies on random sampling of $z$ in the first stage to cover the state-space. However, given the structure of the epigraph form constrained OCP, it should be possible to perform both optimizations simultaneously such that only a single stage of optimization is required.
Moreover, the current method does not account for model errors and parametric uncertainties which may compromise the safety and stability of the learned controllers. An extension of the current method to consider this would allow for more robustness when deploying such a controller to the hardware systems in the real world. Additionally, PPO is an on-policy online reinforcement learning algorithm. Consequently, a simulator of the dynamics is necessary to solve the EF-COCP problem. Extending this work to off-policy offline reinforcement learning setting will allow our method to be applied to settings when a dynamics simulator is not available.
Finally, it is difficult to provide useful statements about the convergence of practical DeepRL algorithms. Nevertheless, it is important to understand properties of the learned policy, especially as violations of safety constraints are serious and undesirable in any system. We leave this as future work. 
\FloatBarrier

\section*{Acknowledgments}
This work is partially supported by the MIT Lincoln Lab under the Safety in Aerobatic Flight Regimes (SAFR) program. However, this article solely reflects the opinions and conclusions of its authors and not the MIT Lincoln Lab.

\bibliographystyle{plainnat}
\bibliography{references}

\newpage
\onecolumn

\begin{appendices}
\numberwithin{equation}{section}
\renewcommand{\theequation}{\thesection.\arabic{equation}}
\renewcommand{\thesubsection}{\thesection\arabic{subsection}}
\renewcommand{\thesubsectiondis}{\thesubsection}

\crefalias{section}{appendix}
\crefalias{subsection}{appendix}

\setlength{\parindent}{0pt}
\setlength{\parskip}{2pt plus1pt minus0.5pt}

\newpage
\section{Proofs} \label{app:sec:proofs}

\subsection{Proof that \texorpdfstring{$V^{l, \pi}$}{V\^{}\{l,π\}} is a discrete-time Lyapunov function} \label{app:sec:proof:discr_lyap}
Before we begin the statement of the theorem and its proof, define $\mathcal{K}$ to be the class of functions $\gamma : \Rb_{\geq0} \to \Rb_{\geq 0}$ that is continuous, zero at zero and strictly increasing. Let $\mathcal{K}_{\infty}$ to be the class that are $\mathcal{K}$ and are also unbounded.
We then have the following definition of a Lyapunov function for discrete-time systems, which we modify from \cite[Ch.2]{grune2017nonlinear} to use $\sigma(\cdot)$ instead of $\norm{\cdot}_{x^{\text{ref}}}$.

\begin{mdframed}[style=ThmFrame]
\begin{restatable}[Discrete-Time Lyapunov Function]{defi}{}\label{app:thm:lya_def}
    Suppose the system has discrete-time dynamics
    \begin{equation}
        x_{k+1} = f(x_k, u_k),
    \end{equation}
    for states $x_k \in \XSet$ and controls $u_k \in \USet$.
    Consider a reference set $\GoalSet$ and a subset of the state space $\ReachSet \subseteq \XSet$.
    Let $\sigma : \XSet \to \Rb_{\geq 0}$ be a state measure (as in \cite{grimm2005model}) that is continuous and positive-definite.
    A function $V : \ReachSet \to \Rb_{\geq0}$ is a \textit{uniform Lyapunov function} on $\ReachSet$ if the following conditions are satisfied.
    \begin{enumerate}[label=(\roman*)]
        \item There exist functions $\alpha_1, \alpha_2 \in \mathcal{K}_\infty$ such that
        \begin{equation}
            \alpha_1( \sigma(x) ) \leq V(x) \leq \alpha_2( \sigma(x) )
        \end{equation}
        holds for all $x \in \ReachSet$.
\item There exists a function $\alpha_V \in \mathcal{K}$ such that
        \begin{equation}
            V(x_{k+1}) \leq V(x_k) - \alpha_V( \sigma(x) )
        \end{equation}
        holds for all $x_k \in \ReachSet$.
    \end{enumerate}
\end{restatable}
\end{mdframed}

\begin{mdframed}[style=ThmFrame]
\begin{restatable}[Policy Value Function is Lyapunov]{thm}{}\label{app:thm:pol_val_lya}
    Let $\pi : \XSet \to \USet$ be an arbitrary deterministic policy, and define $V^{l, \pi} : \XSet \to \Rb_{\geq 0} \cup \{ +\infty \}$ to be the policy value function
    \begin{equation} \label{eq:polval:V_def}
        V^{l, \pi}(x_0) \coloneqq \sum_{k=0}^{\infty} l(x_k), \quad x_{k+1} = f(x_k, \pi(x_k))
    \end{equation}
    for cost function $l : \XSet \to \Rb_{\geq 0}$ and discrete dynamics $f:\XSet \times \USet \to \XSet$. Let $\FiniteSet$ denote the set where $V^{l, \pi}$ is finite, i.e.,
    \begin{equation}
        \FiniteSet \coloneqq \Set{ x | V^{l, \pi}(x) < \infty },
    \end{equation}
    and let $\sigma : \XSet \to \Rb_{\geq 0}$ be a state measure (as in \cite{grimm2005model}) that is continuous and positive-definite.
    Suppose that the following holds for the cost function $l$ and the policy value function $V^{l, \pi}$.
    \begin{enumerate}[label=(\roman*)]
        \item There exists $\overline{\alpha} \in \mathcal{K}_{\infty}$ such that, for any $x \in \FiniteSet$,
        \begin{equation} \label{eq:polval:V_ub}
            V^{l, \pi}(x) \leq \overline{\alpha}( \sigma(x) )
        \end{equation}
        \item There exists $\overline{\rho} \in \mathcal{K}_{\infty}$ such that, for any $x \in \FiniteSet$,
        \begin{equation} \label{eq:polval:l_lb}
            l(x) \geq \overline{\rho}(\sigma(x))
        \end{equation}
    \end{enumerate}
    Then, $V^{l, \pi}$ is a Lyapunov function on $\FiniteSet$.
\end{restatable}
\end{mdframed}

\begin{proof}
    By the definition of $V^{l, \pi}$, use of dynamic programming shows that
    \begin{equation} \label{proof:polval:dp}
        V^{l, \pi}(x_k) = l(x_k) + V^{l, \pi}( x_{k+1} ).
    \end{equation}
    Since $V^{l, \pi} \geq 0$ by definition \eqref{eq:polval:V_def}, by using \eqref{eq:polval:l_lb} and \eqref{proof:polval:dp} we can conservatively lower bound $V^{l, \pi}$ in terms of $\sigma$ on $\FiniteSet$ as 
    \begin{equation} \label{proof:polval:V_lb}
        V^{l, \pi}(x) \geq l(x) \geq \overline{\rho}(\sigma(x)).
    \end{equation}
    Combining the same two equations again without dropping $V^{l, \pi}(x_{k+1})$, we can also show that for $x \in \FiniteSet$,
    \begin{equation} \label{proof:polval:V_descent}
        V^{l, \pi}( x_{k+1} ) = V^{l, \pi}(x_k) - l(x_k) \leq V^{l, \pi}(x_k) - \overline{\rho}(\sigma(x)).
    \end{equation}
    Combining \eqref{eq:polval:V_ub}, \eqref{proof:polval:V_lb} and \eqref{proof:polval:V_descent} then gives us that for $x_k \in \FiniteSet$,
    \begin{subequations}
    \begin{align}
        \overline{\rho}(\sigma(x)) &\leq V^{l, \pi}(x) \leq \overline{\alpha}( \sigma(x) ), \label{proof:polval:V_bounds} \\
        V^{l, \pi}(x_{k+1}) &\leq V^{l, \pi}(x_k) - \overline{\rho}( \sigma(x) ) \label{proof:polval:V_descent2}.
    \end{align}
    Since $\overline{\rho} \in \mathcal{K}_{\infty}$, $\overline{\alpha} \in \mathcal{K}_{\infty}$,
    \eqref{proof:polval:V_bounds} and \eqref{proof:polval:V_descent2} thus show that $V^{l, \pi}$ is a Lyapunov function on $\FiniteSet$ by \Cref{app:thm:lya_def}.
    \end{subequations}
\end{proof}

\vspace{\baselineskip}

From \Cref{app:thm:pol_val_lya}, we can then apply the standard proof of local asymptotic stability using Lyapunov functions \cite{grune2017nonlinear} to show asymptotic stability.

\begin{mdframed}[style=ThmFrame]
\begin{restatable}[]{coroll}{}\label{app:thm:pol_stable}
    Define the set $\ZeroSet$ to be the zero-set of $V^{l, \pi}$, i.e.,
    \begin{equation}
        \ZeroSet \coloneqq \Set{ x | V^{l, \pi}(x) = 0 }.
    \end{equation}
    Then, $\ZeroSet$ is also the zero-set of $\sigma$, i.e.,
    \begin{equation}
        \ZeroSet = \Set{ x | \sigma(x) = 0 }.
    \end{equation}
    Moreover, $\ZeroSet$ is locally asymptotically stable within $\FiniteSet$ under the controller $\pi$ on $\FiniteSet$.
\end{restatable}
\end{mdframed}

\begin{proof}
    First, note that by the definition of $\mathcal{K}$, \eqref{proof:polval:V_lb} implies that for $x \in \ZeroSet$,
    \begin{equation}
        0 \leq \overline{\rho}(x) \leq V^{l, \pi}(x) = 0.
    \end{equation}
    Moreover, since $\overline{\rho}$ is strictly increasing,
    \begin{equation}
        \overline{\rho}(x) > 0 \implies V^{l, \pi}(x) > 0.
    \end{equation}
    Hence, the zero-set $\ZeroSet$ of $V^{l, \pi}$ is also the zero-set of $\sigma$.
    Applying Theorem 2.19 from \cite{grune2017nonlinear} using the policy value function as the Lyapunov function as shown in \Cref{app:thm:pol_val_lya} then gives us the result.
\end{proof}

\begin{mdframed}[style=ThmFrame]
\begin{restatable}[]{note}{}
    As noted in the main paper, while \Cref{app:thm:pol_val_lya} and \Cref{app:thm:pol_stable} show that we can use $V^{l, \pi}$ to show stability for \textit{any} policy $\pi$ within the region $\mathcal{F}$ under assumptions \eqref{eq:polval:l_lb} and \eqref{eq:polval:V_ub}, we note that $\mathcal{F}$ may be a tiny set or even empty. Hence, the theorems above do not give us a direct method of constructing stable controllers. Nevertheless, the above theorems provide intuition on the relationship between the optimality of a policy (measured by the size of $\mathcal{F}$) and its stability, which we use when solving the infinite-horizon constrained OCP in the main paper.
\end{restatable}
\end{mdframed} \newpage

\ifarxiv
    \subsection{Equivalence of \eqrefApp{eq:opt:ex:linear_obj} and \eqrefApp{eq:opt:ex:linear_obj2}}
\else
    \subsection{Equivalence of (11) and (12)} 
\fi
\label{app:sec:proof:minimax}

\begin{mdframed}[style=ThmFrame]
\begin{restatable}[]{thm}{}\label{app:thm:pol_val_lya}
    Let $x \in \Rb^n$ and $z \in \Rb$, and let $g : \Rb^n \times \Rb \to \Rb$ be a continuous (potentially non-differentiable) function.
    Then, if a solution exists (i.e., an optimal $x^*, z^*$ exist, are finite), then the following optimization problems are equivalent.

    \vspace{1ex}
    \begin{minipage}{0.495\linewidth}
        \begin{mini}[2]
        {x,z}{z\hphantom{acbdefghijkl}}{\label{eq:reform_proof:L}}{}
        \addConstraint{g(x, z)}{\leq 0,}
        \end{mini}
    \end{minipage}
    \hfill
    \begin{minipage}{0.495\linewidth}
        \begin{mini}[2]
        {z}{z\hphantom{acbdefghijkl}}{\label{eq:reform_proof:R}}{}
        \addConstraint{\Big[ \min_x g(x, z) \Big]}{\leq 0,}
        \end{mini}
    \end{minipage}
\end{restatable}
\end{mdframed}

\begin{proof}
    We begin by comparing the Lagrangian primal problem of \cref{eq:reform_proof:L} and \cref{eq:reform_proof:R}.

    \begin{minipage}{0.495\linewidth}
        \begin{equation} \label{eq:reform_proof:lagr_primal:L}
            \begin{split}
            &\mathrel{\phantom{=}} \min_z \min_x \max_{\lambda \geq 0} z + \lambda \, g(x, z) \\
            &= \min_z \left\{ z + \min_x \max_{\lambda \geq 0} \lambda \, g(x, z) \right\}
            \end{split}
        \end{equation}
    \end{minipage}
    \hfill
    \begin{minipage}{0.495\linewidth}
        \begin{equation} \label{eq:reform_proof:lagr_primal:R}
            \begin{split}
            &\mathrel{\phantom{=}} \min_z \max_{\lambda \geq 0} z + \lambda \Big[ \min_x g(x, z) \Big] \\
            &= \min_z \left\{ z + \max_{\lambda \geq 0} \min_x \lambda \, g(x, z) \right\}
            \end{split}
        \end{equation}
    \end{minipage}

    Comparing the two, the only difference is that the order of $\min_x$ and $\max_{\lambda}$ are flipped. Hence, it is sufficient to show that,
    for any $z$ where $\min_x g(x, z) < 0$,
    \begin{equation}
        p^* \coloneqq \min_x \max_{\lambda \geq 0} \lambda \, g(x, z) = \max_{\lambda \geq 0} \min_x \lambda \, g(x, z) \eqqcolon d^*.
    \end{equation}
    Note that this is exactly equivalent to showing that strong duality holds for the following constraint satisfaction problem.
    \begin{mini}[2]
    {x}{0}{\label{eq:reform_proof:constr_satis}}{}
    \addConstraint{g(x, z)}{\leq 0,}
    \end{mini}
    We now prove that strong duality holds for the above problem in a similar fashion to the proof that Slater's condition is a sufficient condition for strong duality to hold in convex optimization problems \cite{boyd2004convex}.

    Define the set $\mathcal{A} \subseteq \Rb^n \times \Rb$ as
    \begin{align}
        \mathcal{A}
        &\coloneqq \Set{(u, t) | \exists x, \; g(x, z) \leq u, \quad 0 \leq t}, \\
        &= \Set{u | \inf_x g(x, z) \leq u} \times [0, \infty).
    \end{align}
    Note that $\mathcal{A}$ is convex. Furthermore, since a feasible solution exists by assumption, we have that
    \begin{equation}
        p^* = \min_x \max_{\lambda\geq0} \lambda \, g(x, z) = \min_x \begin{dcases} \infty & g(x, z) > 0 \\ 0 & g(x, z) \leq 0 \end{dcases} \; = 0.
    \end{equation}
    We now define a second set $\mathcal{B} \subseteq \Rb^n \times \Rb$ as
    \begin{align}
        \mathcal{B}
        &\coloneqq \Set{(0, s) | s < p^*}, \\
        &= \{ 0 \} \times (-\infty, 0).
    \end{align}
    Note that $\mathcal{B}$ is also convex, and that the sets $\mathcal{A}$ and $\mathcal{B}$ do not intersect. We can then invoke the separating hyperplane theorem to show that there exists a $(\tilde{\lambda}, \mu) \not= 0$ and a $\alpha$ that defines a hyperplane which separates the two sets, i.e.,
    \begin{align}
        (u, t) \in \mathcal{A} \implies \tilde{\lambda} u + \mu t \geq \alpha \label{eq:reform_proof:sep_A} \\
        (u, s) \in \mathcal{B} \implies \tilde{\lambda} u + \mu s \leq \alpha \label{eq:reform_proof:sep_B}
    \end{align}
    In \eqref{eq:reform_proof:sep_A}, since both $u$ and $t$ are unbounded above, we must have $\tilde{\lambda} \geq 0$ and $\mu \geq 0$. 
    Furthermore, in \eqref{eq:reform_proof:sep_B}, since $s < p^*$, we have that $\mu p^* \leq \alpha$. Combining both then gives us that for all $x$,
    \begin{equation}
0 = p^* = \mu p^* \leq \alpha \leq \tilde{\lambda} g(x, z).
    \end{equation}
    Minimizing the RHS over $x$ then maximizing over $\tilde{\lambda}$ then gives us that
    \begin{equation}
        p^* \leq \min_x \tilde{\lambda} g(x, z) \leq \max_{\lambda} \min_x \lambda g(x, z) = d^*.
    \end{equation}
    Finally, by weak duality, we have that
    \begin{equation}
        p^* \geq d^*.
    \end{equation}
    Combining the two then allows us to conclude that $p^* = d^*$.
\end{proof} 

\newpage
\section{Understanding the role of \texorpdfstring{$z$}{z} in the EFCOCP inner problem} \label{app:sec:z_role}
In this section, we provide more intuition about the role of $z$ on the learned policy $\pi$ and the learned value function $V^{\pi}$. We first restate the EFCOCP inner problem below.
\begin{equation}
    \tilde{J}^{\pi}(x_0, z) \coloneqq \max\left\{ \max_{k \geq 0} h(x_k), \sum_{k=0}^\infty l(x_k) - z \right\}.
\end{equation}
As $z \to -\infty$, the cost (i.e., stability) related term dominates the $\max$.
Consequently, we should see that the optimal policy will prioritize stability. On the other hand, as $z \to \infty$, the constraint (i.e., safety) related term dominates the $\max$.
In this case, the optimal policy will prioritize safety.
Moreover, if the optimal policy is safe under the unconstrained minimizer, i.e.,
\begin{equation}
    h_{\max} \coloneqq \max_{k \geq 0} h(x_k) \leq 0,
\end{equation}
then the second term will be larger than the first. Since the second term is non-negative, we have that for any $z \in (-h_{\max}, 0]$,
\begin{equation}
    \sum_{k=0}^\infty l(x_k) - z \geq -z > h_{\max}
\end{equation}
Consequently, for such a choice of $z$,
\begin{align}
    \tilde{J}^{\pi}(x_0, z),
    &= \max\left\{ \max_{k \geq 0} h(x_k), \sum_{k=0}^\infty l(x_k) - z\right\}, \\
    &= \max\left\{ h_{\max}, \sum_{k=0}^\infty l(x_k) - z \right\}, \\
    &= \sum_{k=0}^\infty l(x_k),
\end{align}
and we recover the unconstrained optimizer.

\begin{figure}[H]
    \centering
    \includegraphics[width=0.87\linewidth]{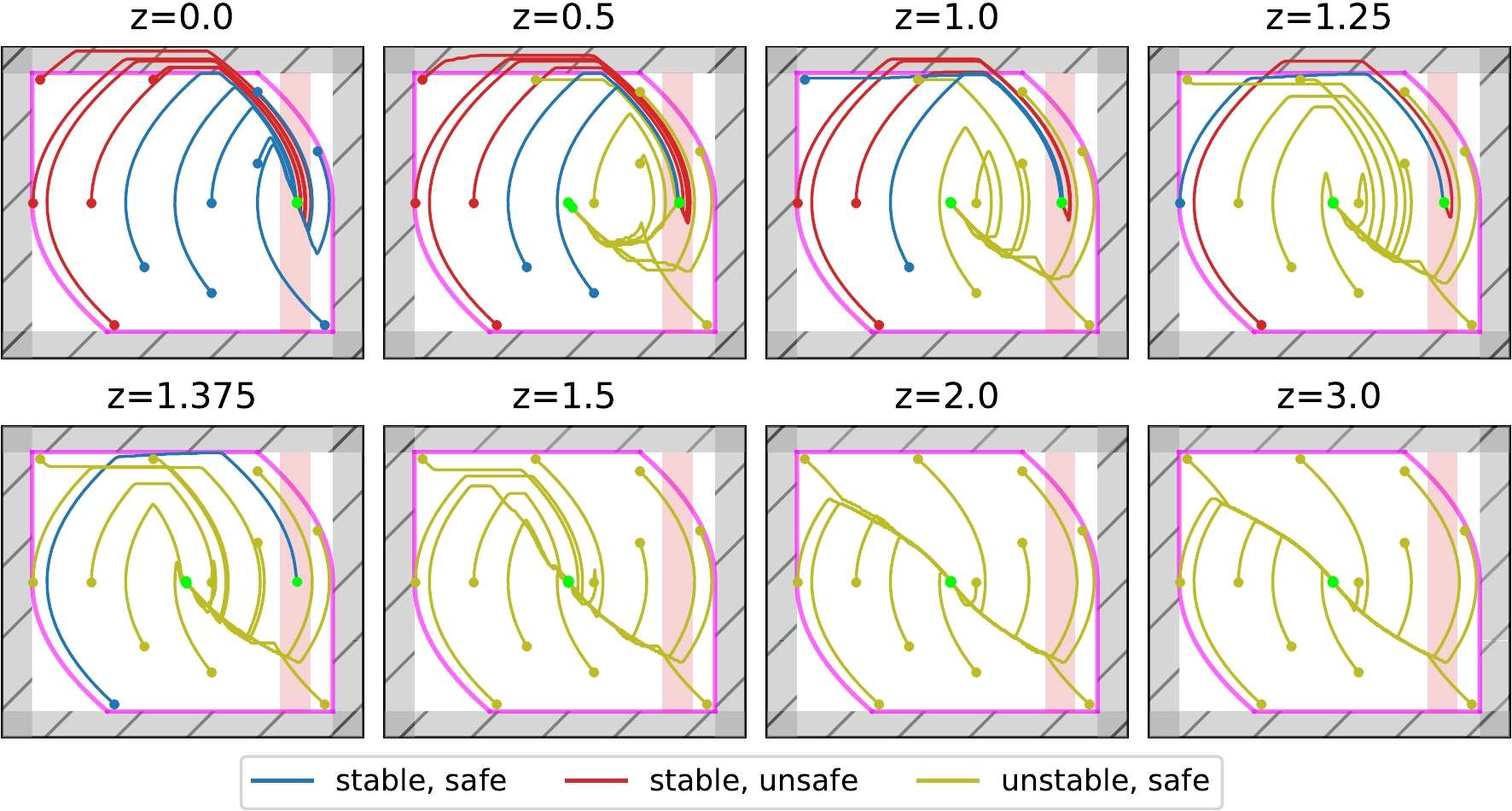}
    \caption{Comparison of the policy $\pi(\cdot, z)$ for different values of $z$ on the 1D double integrator.}
    \label{fig:app:2dint:z_sens}
\end{figure}

We now compare the policy rollouts for different values of $z$ on different systems.
The policy rollouts for the 1D double-integrator system are shown in \Cref{fig:app:2dint:z_sens}.
Note that for $z=0$, states whose unconstrained minimizer are safe follow the unconstrained optimal trajectory and converge to the goal region.
As $z$ increases, the policy focuses more on constraint satisfaction.
Consequently, trajectories that were originally unsafe (\textcolor{red}{red}) become safe (\textcolor{blue}{blue}).
However, as $z$ increases further, the policy focuses too much on minimizing constraint function (i.e., $\max_{k \geq 0} h(x_k)$) and converges to the minimizer of $h$ instead of the goal region (\textcolor{olive}{olive}). 

\begin{figure}
    \centering
    \begin{subfigure}[b]{0.32\linewidth}
        \centering
        \includegraphics[width=\linewidth]{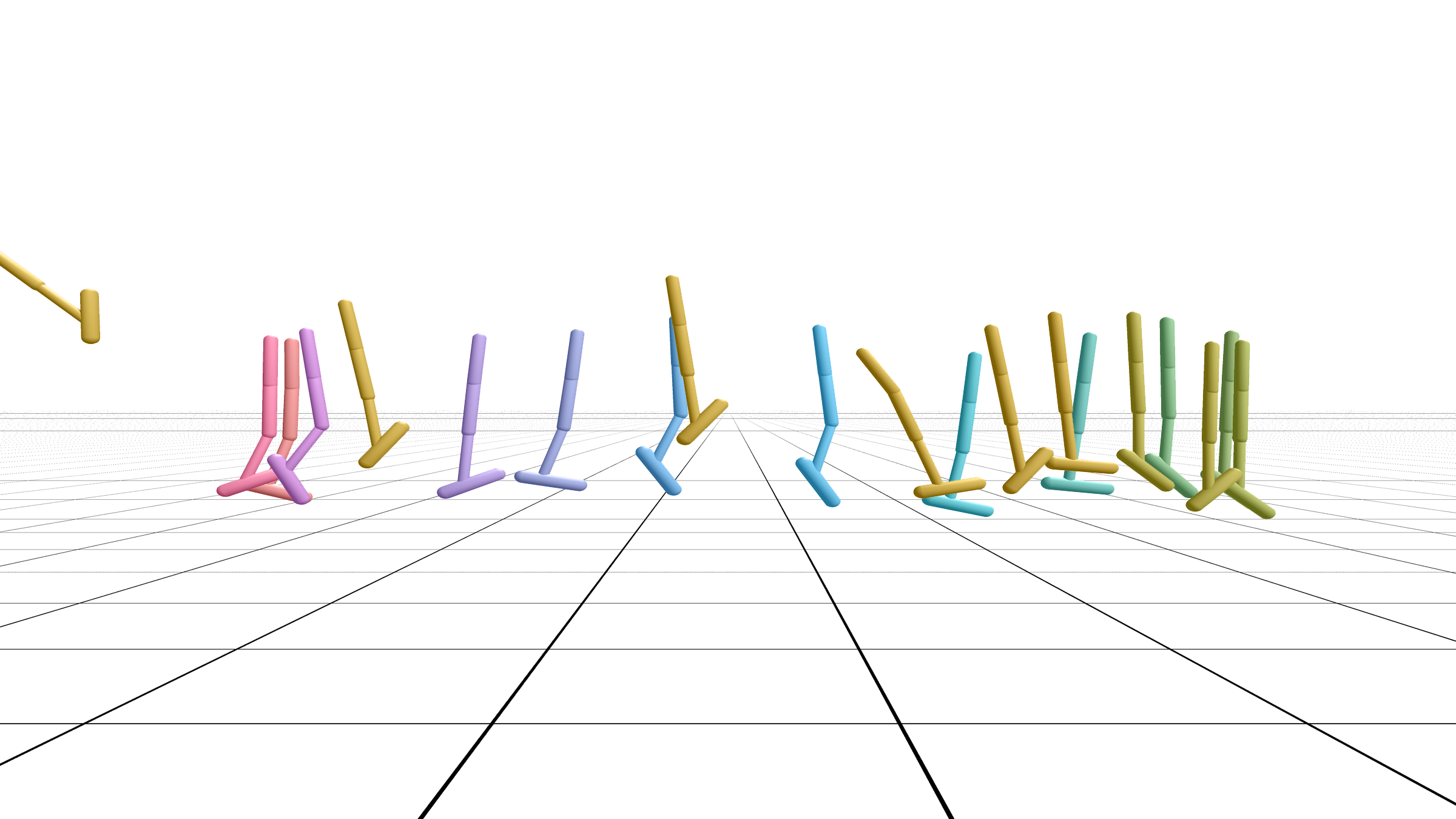}
        \includegraphics[width=\linewidth]{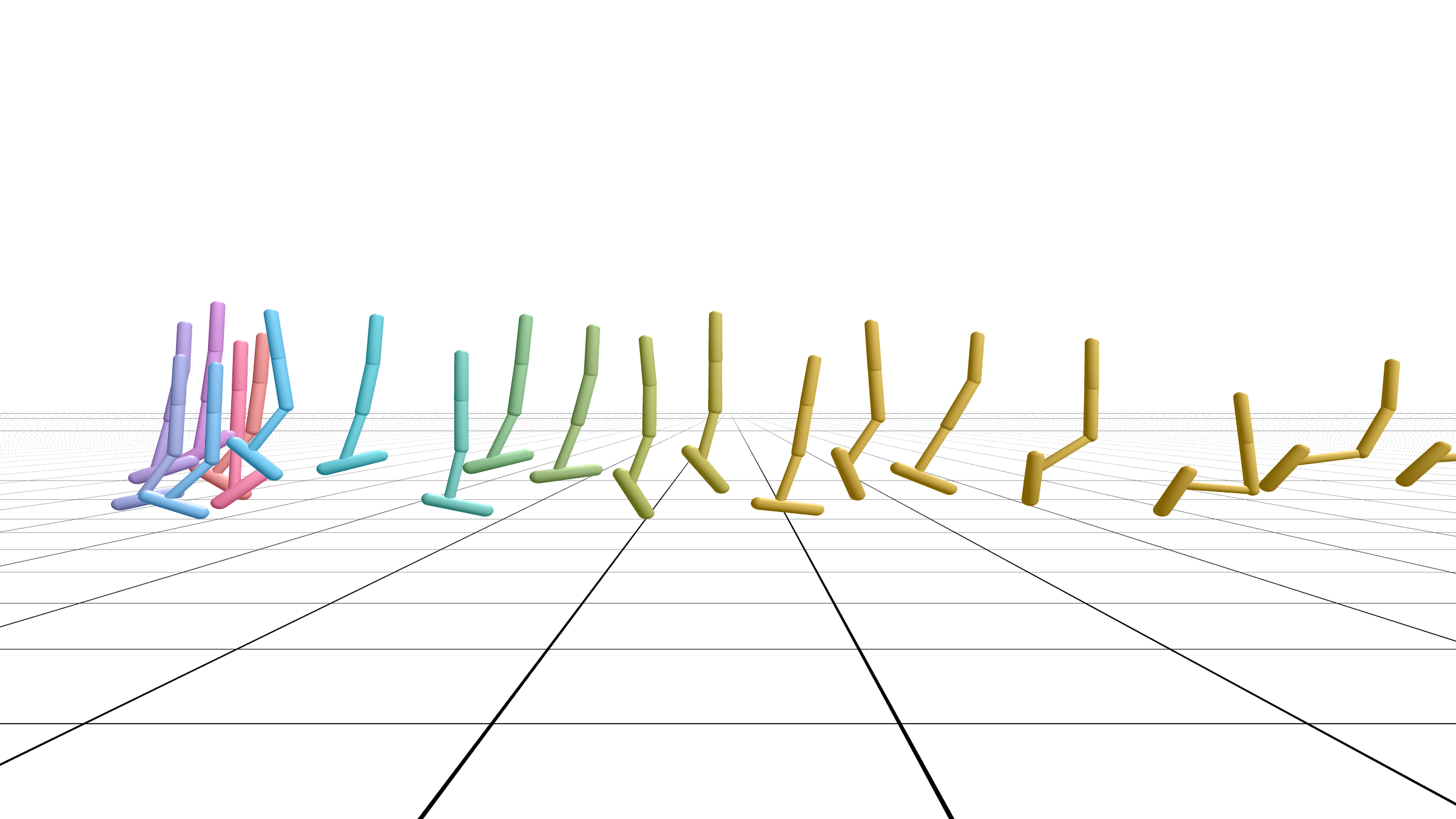}
        \includegraphics[width=\linewidth]{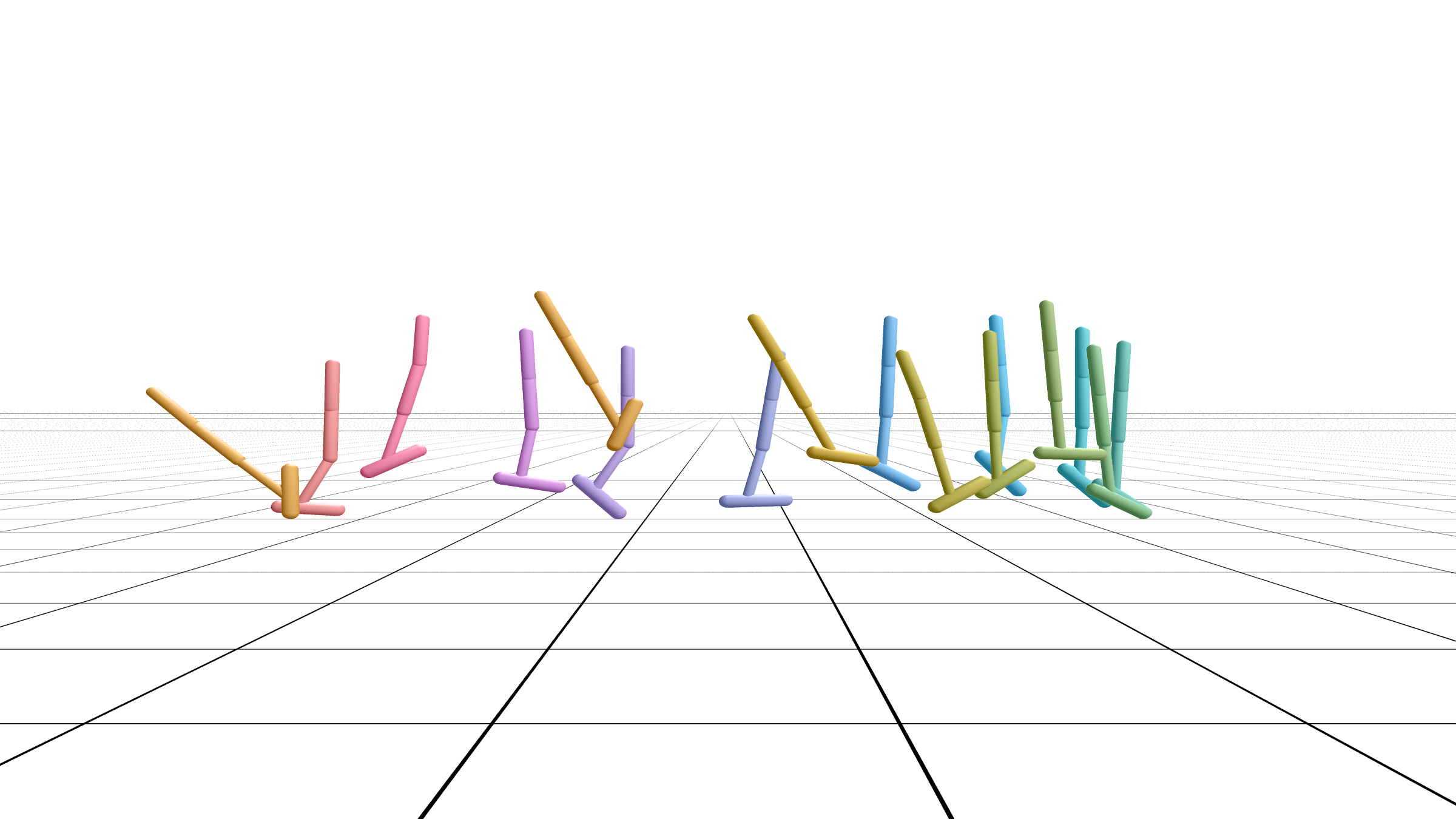}
        \caption{$z=0.0$}
    \end{subfigure}
    \hfill
    \begin{subfigure}[b]{0.32\linewidth}
        \centering
        \includegraphics[width=\linewidth]{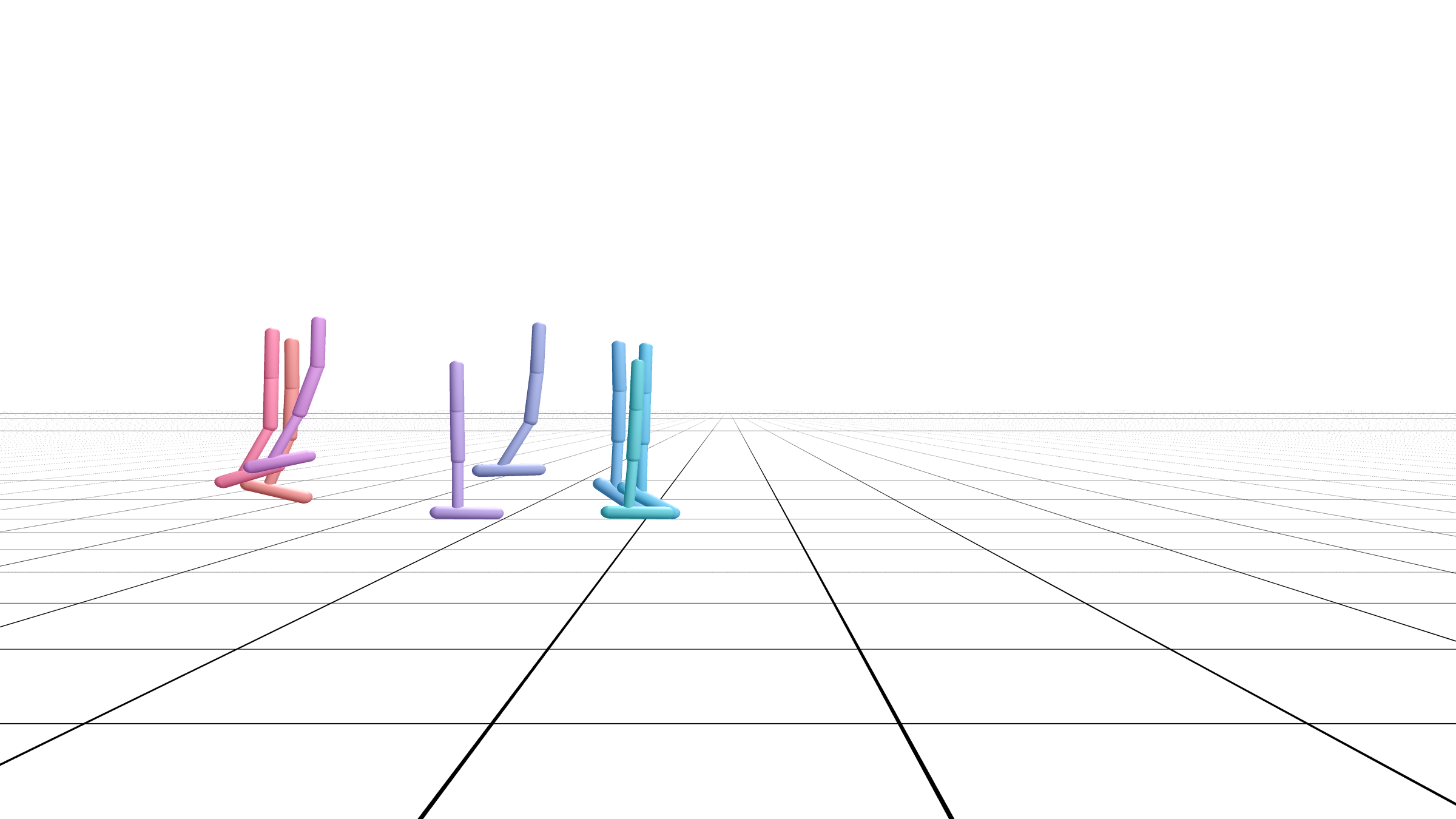}
        \includegraphics[width=\linewidth]{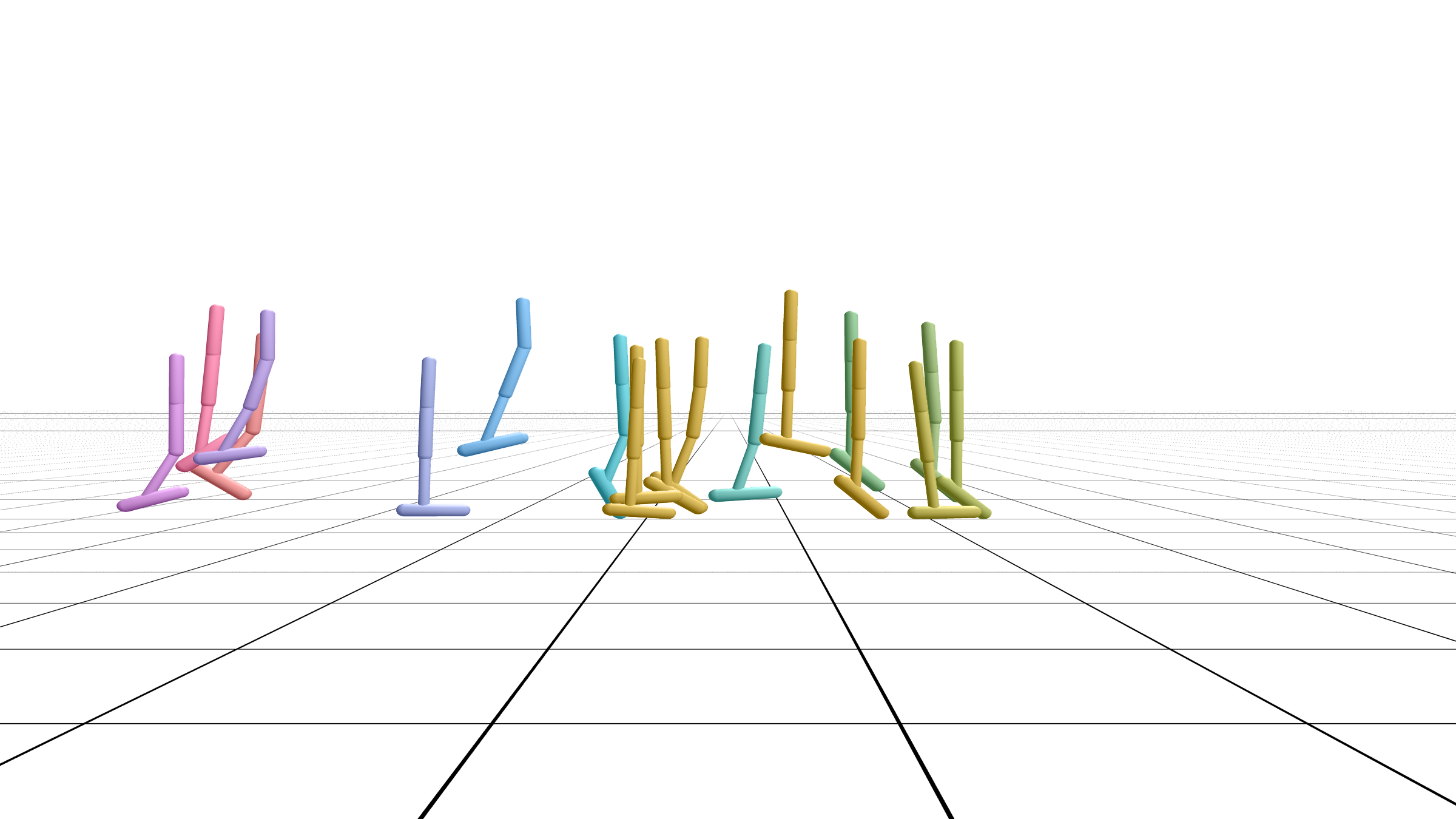}
        \includegraphics[width=\linewidth]{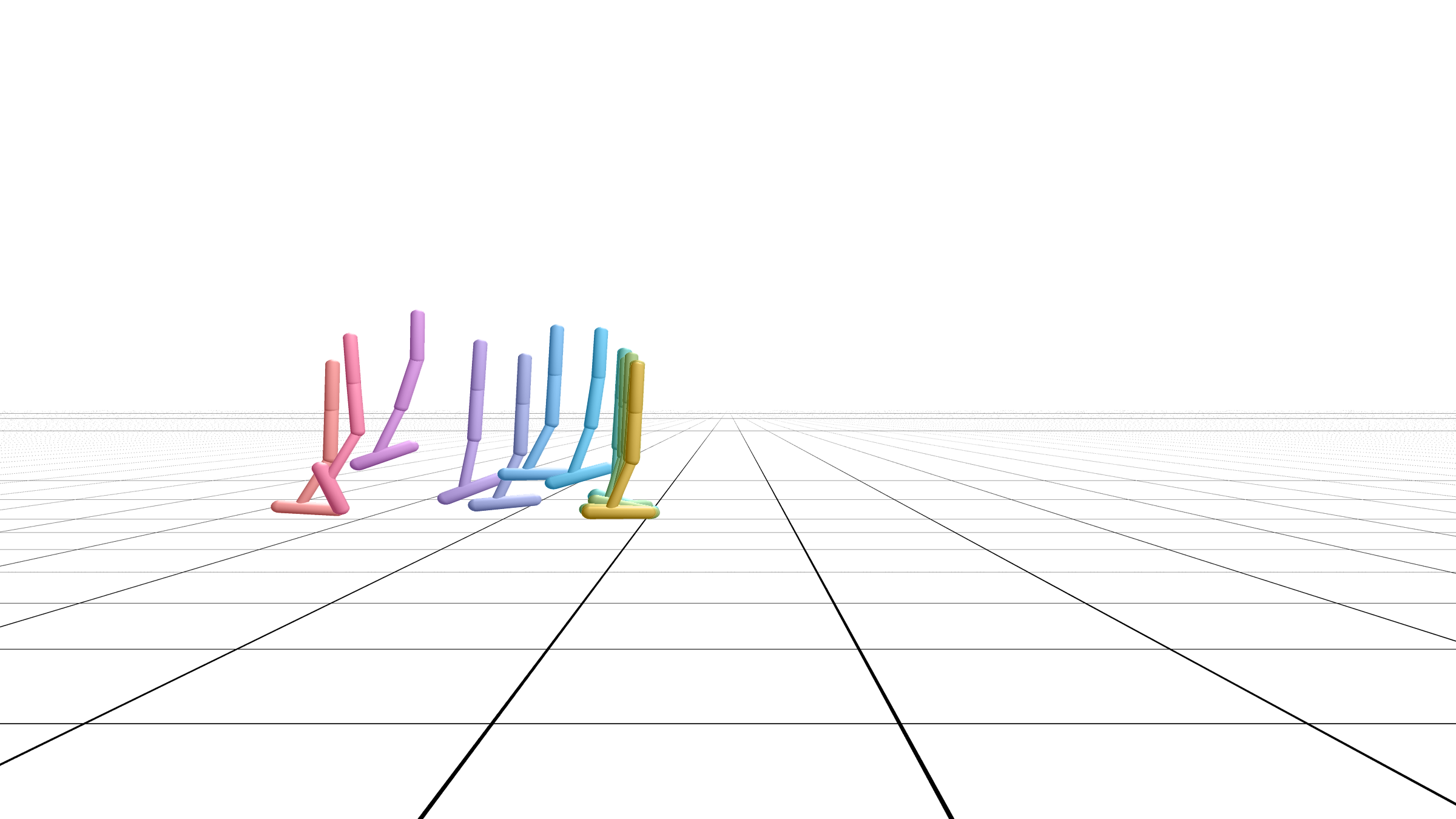}
        \caption{$z=0.75$}
    \end{subfigure}
    \hfill
    \begin{subfigure}[b]{0.32\linewidth}
        \centering
        \includegraphics[width=\linewidth]{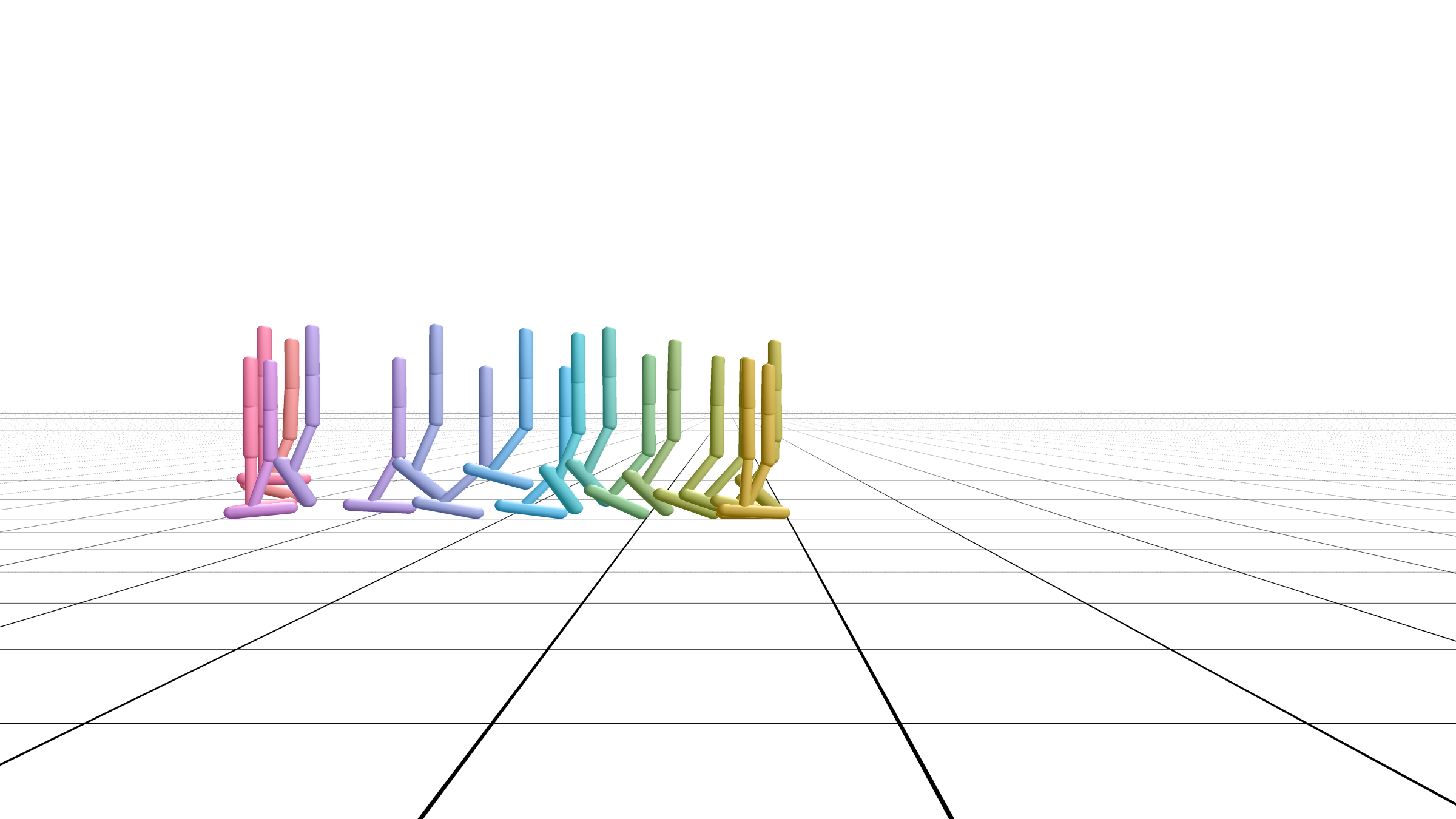}
        \includegraphics[width=\linewidth]{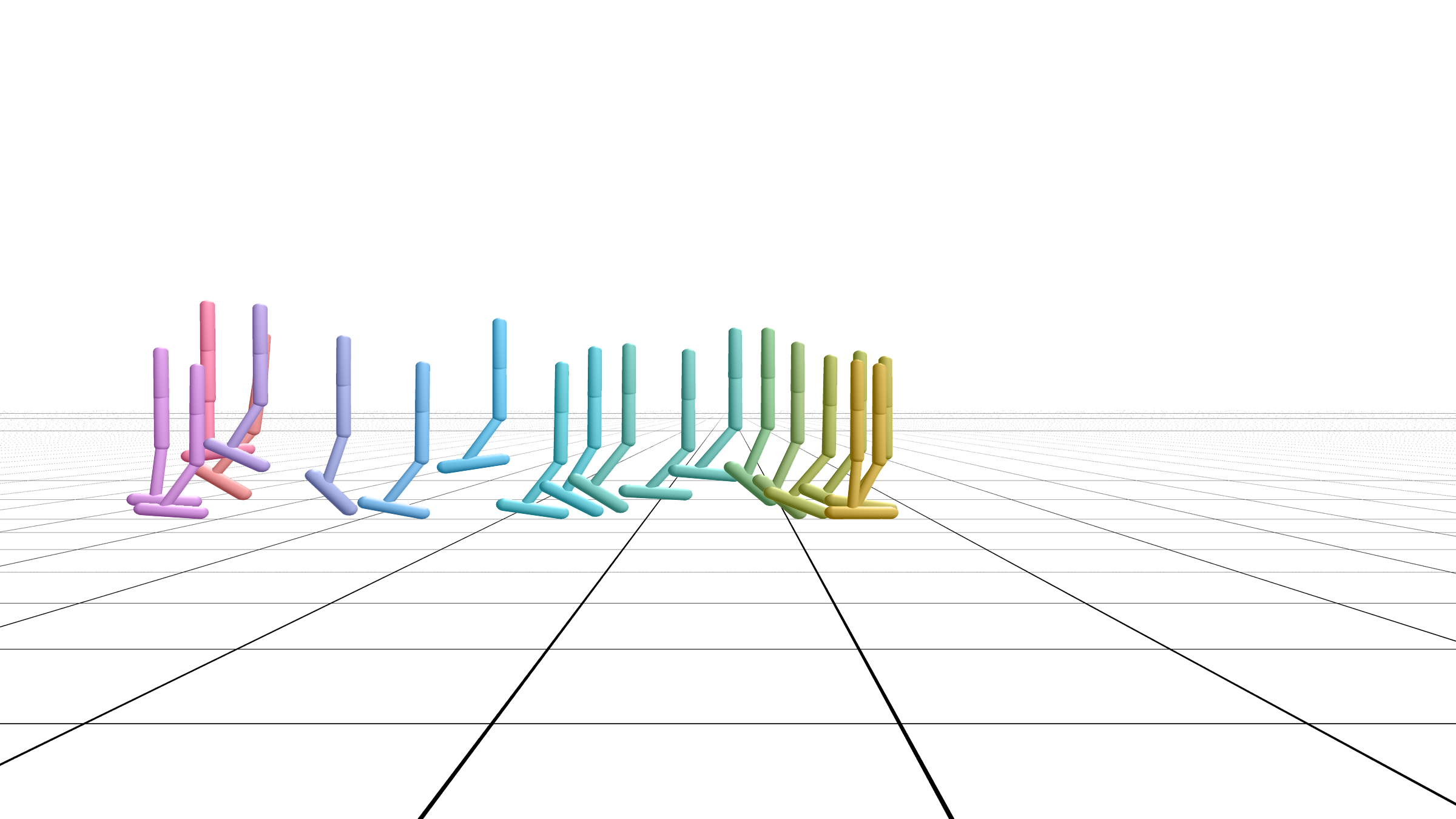}
        \includegraphics[width=\linewidth]{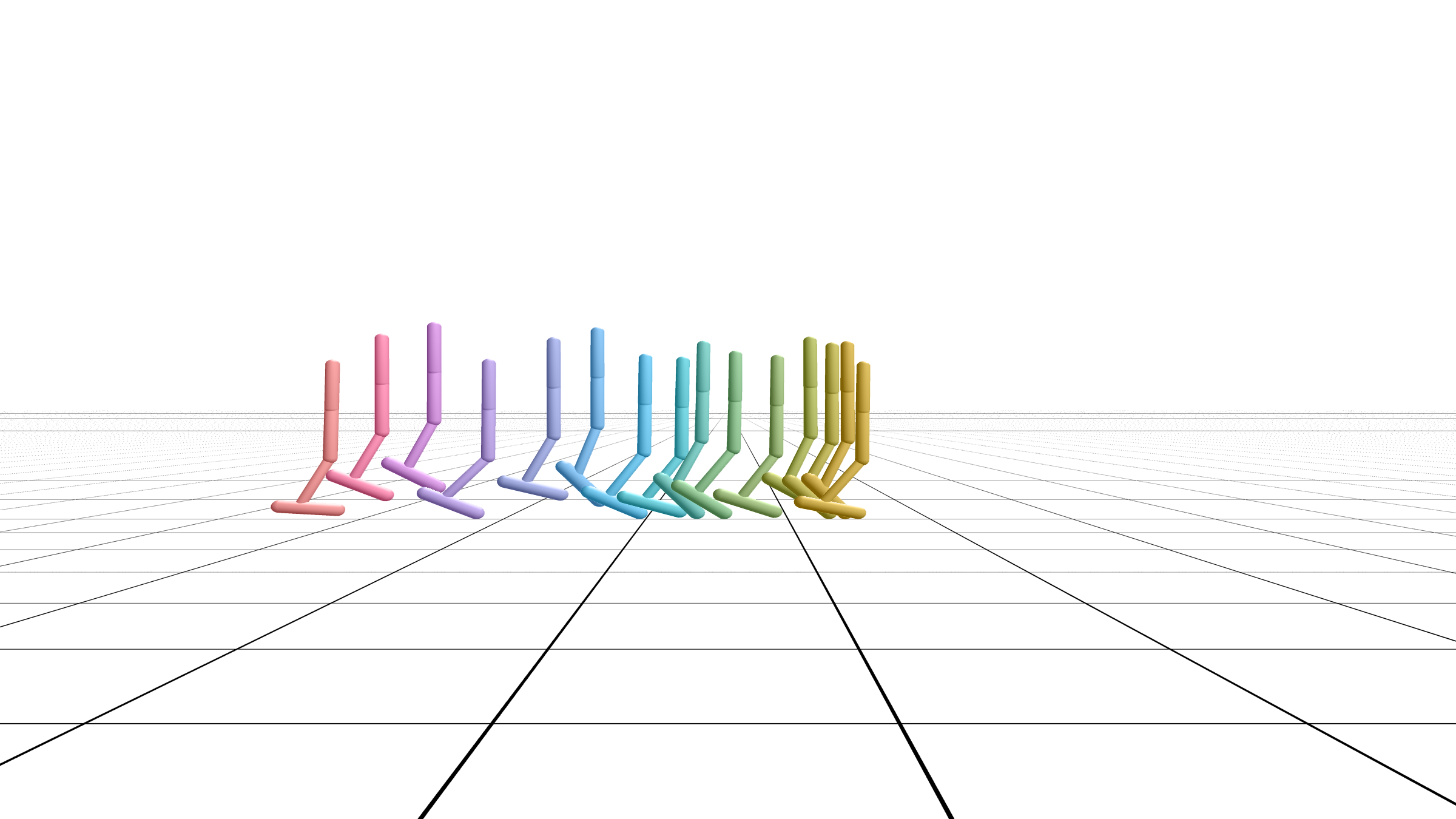}
        \caption{$z=2.8$}
    \end{subfigure}
    \caption{Comparison of the policy $\pi(\cdot, z)$ for $z=0$, $z=0.75$ and $z=2.8$ respectively from left to right on the Hopper system. The direction of time follows the colors \textcolor{red}{red}, \textcolor{purple}{purple}, \textcolor{blue}{blue}, \textcolor{green}{green}, \textcolor{MaterialYellow600}{yellow}. Each row represents a different initial condition.
    The policy for $z=0.0$ is too aggressive and eventually topples, violating the safety constraints. In contrast, the policy for $z=2.8$ prioritizes safety by keeping the torso vertical, but hops very slowly in doing so and also overshoots the goal region (compare with $z=2.8$). Taking $z$ to be a value between these two extremes (e.g., $z=0.75$) stabilizes to the goal while maintaining safety. By training a policy $\pi$ that is conditioned on $z$, we can maintain safety and obtain a stabilizing controller despite $\pi$ being suboptimal by learning a proper value of $z^*$.}
    \label{fig:app:hopper:z_sens}
\end{figure} We next show the policy rollouts on the Hopper system for different values of $z$ in \Cref{fig:app:hopper:z_sens}. Again, we can see that larger values of $z$ correlates to higher emphasis on safety. In the case of Hopper, note that the optimal unconstrained optimizer should be able to maintain safety. However, despite the learned policy being suboptimal (and hence unsafe), we are still able to obtain a safe final policy by using $z > 0$. 
\newpage
\section{Discounting in EFPPO} \label{app:sec:discounts}
As noted in the main text, we randomly sample $z$ from $[z, z_{\max}]$ when solving the inner problem of EFCOCP, where $z_{\max}$ is an upper bound of the total cost under the optimal policy $\pi^*$, i.e.,
\begin{equation}
    z_{\max} \geq \sum_{k=0}^\infty l(x_k),
\end{equation}
for any trajectory $\{ x_k \}_{k=0}^\infty$.
However, if the system under the optimal policy does not stabilize to the zero-set of $l$ fast enough (or not at all) due to lack of controllability, then this may be infinite.
While this is not a problem for the solution of the optimization problem if the system is not controllable from $x_0$, it is problematic when we apply reinforcement learning to the problem and learn a neural network that approximates the policy value function $V^{\pi}$. Such a term will dominate the loss function when training $V^{\pi}$.

To alleviate this, we apply a small discount factor $\gamma \in (0, 1)$, taken to be $0.97$ in all of our experiments.
Consequently, we now consider the \textit{discounted} EFCOCP inner problem, where the cost function $\tilde{J}^{\pi}$ now takes the form
\begin{equation}
    \tilde{J}(x_0, z) \coloneqq \max\Big\{ \max_{k \geq 0} \gamma^k h(x_k), \sum_{k=0}^\infty \gamma^k l(x_k) - z \Big\}.
\end{equation}
The dynamic programming equations are modified correspondingly, which we derive below.
\begin{align}
    \tilde{V}(x_0, z_0)
    &= \min_{u_{0:\infty}} \max\Big\{ \max_{k \geq 0} \gamma^k h(x_k), \sum_{k=0}^\infty \gamma^k l(x_k) - z \Big\}, \\
    &= \min_{u_{0:\infty}} \max\Big\{ h(x_0), \max_{k \geq 1} \gamma^k h(x_k), \sum_{k=1}^\infty \gamma^k l(x_k) - \big(z - l(x_0) \big) \Big\}, \\
    &= \min_{u_{0}} \max\Big\{ h(x_0),
        \min_{u_{1:\infty}} \max\Big( \max_{k \geq 1} \gamma^k h(x_k), \sum_{k=1}^\infty \gamma^k l(x_k) - \big(z - l(x_0) \big) \Big)
    \Big\}, \\
    &= \min_{u_{0:\infty}} \max\left\{ h(x_0),
        \gamma \min_{u_{1:\infty}} \max\left( \max_{k \geq 0} \gamma^k h(x_{k+1}), \sum_{k=0}^\infty \gamma^k l(x_{k+1}) - \frac{z - l(x_0)}{\gamma} \right)
    \right\}, \\
    &= \min_{u_{0:\infty}} \max\left\{ h(x_0),
        \gamma V\left( x_1, \frac{z - l(x_0)}{\gamma} \right)
    \right\}, \\
    &= \min_{u_{0:\infty}} \max\left\{ h(x_0), \gamma V\left( x_1, z_{1} \right)
    \right\},
\end{align}
where the ``dynnamics'' for $z$ now read
\begin{equation}
    z_{k+1} = \frac{z_k - l(x_k)}{\gamma}.
\end{equation}
Following this, the policy value function $V^{\pi}$ and policy action-value function $Q^{\pi}$ used for EFPPO are modified accordingly.

With the discounted formulation, $V^\pi$ is now finite assuming $h(x_k)$ does not explode and $l(x_k)$ does not grow faster than $\gamma^k$, which is satisfied in most practical problems where the system has enough control authority. While we can find $z_{\max}$ analytically, in practice $z_{\max}$ is found empirically by running the inner loop of EFPPO for several iterations and then taking $z_{\max}$ to be a constant multiple (e.g., $1.5$) of the largest value of $\sum_{k=0}^{\infty} \gamma^k l(x_k)$ seen so far.
Since the initial policy is generally worse than $\pi^*$ (i.e., has larger cost), this procedure yields a conservative over-estimate of $z_{\max}$ that we have found to be robust. In our experiments, this procedure only needs to be performed once for every new task to set $z_{\max}$ and does not require any tuning afterwards.

\newpage
\section{Simulation Details} \label{app:sec:exp_details}
Details for the simulation environments used are provided below.

\subsection{1D Double-Integrator}
\begin{table}[H]
\centering
\begin{booktabs}{
  colspec = {rllrll},
  cell{1}{1,4} = {c=3}{c}, }
\toprule 
    States & _ & _ & Controls & _ & _ \\ \cmidrule[lr]{1-3} \cmidrule[lr]{4-6}
    Index & Symbol & Description & Index & Symbol & Description \\ \midrule
    0 & $p$ & Position & 0 & $a$ & Acceleration \\
    1 & $v$ & Velocity & \\
    \bottomrule
\end{booktabs}
\caption{States and Controls for the 1D Double-Integrator}
\label{tab:2dint_overview}
\end{table}

The 1D Double-Integrator is a system with $2$ state and $1$ control dimensions (see \Cref{tab:2dint_overview}). The dynamics are linear and take the form
\begin{equation}
    \mat{p_{k+1}; v_{k+1}} = \mat{1 \Delta t; 0 1} \mat{p_k; v_k} + \mat{\frac{1}{2}\Delta t^2; \Delta t} \mat{a_k}
\end{equation}
for timestep $\Delta t$. We use $\Delta t = 0.025$.

The control constraints are box constraints within $[-1, 1]$
\begin{equation}
    \abs{a} \leq 1.
\end{equation}
The state constraints (which define the avoid set $\AvoidSet$) are 
\begin{equation}
    \abs{p} \leq 1, \quad \abs{v} \leq 1
\end{equation}
To represent $\AvoidSet$, we define $h(\vx) = \max{h_1(\vx), h_2(\vx)}$, where
\begin{equation}
    h_1(\vx) \coloneqq \abs{p} - 1, \quad h_2(\vx) \coloneqq \abs{v}^3 - 1.
\end{equation}
The goal set $\GoalSet$ is defined as the region
\begin{equation}
    \GoalSet \coloneqq \Set{\vx | p \in [0.65, 0.85]},
\end{equation}
which we represent via the cost function $l$ as
\begin{equation}
    l(x) \coloneqq \big[ \abs{p - 0.75} - 0.1 \big]^+.
\end{equation}

\newpage
\subsection{2D Single-Integrator with Sector Obstacle}
\begin{table}[H]
\centering
\begin{booktabs}{
  colspec = {rllrll},
  cell{1}{1,4} = {c=3}{c}, }
\toprule 
    States & _ & _ & Controls & _ & _ \\ \cmidrule[lr]{1-3} \cmidrule[lr]{4-6}
    Index & Symbol & Description & Index & Symbol & Description \\ \midrule
    0 & $p_x$ & Position along $x$ & 0 & $v_n$ & Velocity along the normal to the origin \\
    1 & $p_y$ & Position along $y$ & 1 & $v_t$ & Velocity along the tangent to the origin\\
    \bottomrule
\end{booktabs}
\caption{States and Controls for the 2D Single-Integrator}
\label{tab:sectorcircle:overview}
\end{table}

The 2D Single-Integrator is a system with $2$ state and $2$ control dimensions (see \Cref{tab:sectorcircle:overview}). The continuous-time dynamics are as
\begin{equation}
    \dv{t} \mat{p_x; p_y} = \frac{1}{\sqrt{p_x^2 + p_y^2}} \mat{-p_x -p_y ; p_y -p_x} \mat{v_n; v_t},
\end{equation}
where the denominator is clipped to prevent division by $0$. The discrete-time dynamics are obtained by discretizing the above using Euler integration with timestep $\Delta t = 0.05$.

The control constraints are box constraints within $[-1, 1]^2$
\begin{equation}
    \abs{v_n} \leq 1, \quad \abs{v_t} \leq 1.
\end{equation}
The state constraints are represented as the set $h(x) = \max(h_0(x), h_1(x))\leq 0$, where 
\begin{equation}
    h_0(\vx) \coloneqq r - 1, \quad
    h_1(\vx) \coloneqq 0.2 (1 - \sqrt{2}) + \sqrt{2} p_x - r.
\end{equation}
where $r \coloneqq \sqrt{p_x^2 + p_y^2}$ denotes the distance to the origin.
$h_0$ defines a circle with radius $1$, while $h_1$ defines the sector obstacle.
The goal set $\GoalSet$ is defined as a circle at the origin with radius $R=0.05$,
which we represent via the cost function
\begin{equation}
    l(\vx) \coloneqq [r - 0.05]^+.
\end{equation}

\newpage
\subsection{Hopper Stabilization}
\begin{table}[H]
\centering
\begin{booktabs}{
  colspec = {rllrll},
  cell{1}{1,4} = {c=3}{c}, }
\toprule 
    States & _ & _ & Controls & _ & _ \\ \cmidrule[lr]{1-3} \cmidrule[lr]{4-6}
    Index & Symbol & Description & Index & Symbol & Description \\ \midrule
     0 & $p_x$      & $x$-coordinate of the torso             & 0 & $\tau_t$ & Torque applied to the thigh motor   \\
     1 & $p_z$      & $z$-coordinate of the torso             & 1 & $\tau_l$ & Torque applied to the leg motor     \\
     2 & $\theta$   & Angle of the torso                      & 2 & $\tau_f$ & Torque applied to the foot motor    \\
     3 & $\theta_t$ & Joint Angle of the thigh                &   &          &                                     \\
     4 & $\theta_l$ & Joint Angle of the leg                  &   &          &                                     \\
     5 & $\theta_f$ & Joint Angle of the foot                 &   &          &                                     \\
     6 & $v_x$      & Velocity of $x$-coordinate of the torso &   &          &                                     \\
     7 & $v_z$      & Velocity of $z$-coordinate of the torso &   &          &                                     \\
     8 & $\omega$   & Angular velocity of the torso           &   &          &                                     \\
     9 & $\omega_t$ & Joint Velocity of the thigh             &   &          &                                     \\
    10 & $\omega_l$ & Joint Velocity of the leg               &   &          &                                     \\
    11 & $\omega_f$ & Joint Velocity of the foot              &   &          &                                     \\
    \bottomrule
\end{booktabs}
\caption{States and Controls for the Hopper}
\label{tab:hopper:overview}
\end{table}

The Hopper is a system with $12$ state and $3$ control dimensions (see \Cref{tab:hopper:overview}) implemented using the Brax \cite{brax2021github} simulator.
Note that the version of Hopper we use includes $p_x$. The original Hopper environment from Brax and other simulators such as Mujoco \cite{todorov2012mujoco} exclude $p_x$, since the goal is to learn a limit-cycle that is independent of $p_x$. In Brax, the dynamics are defined using the rigid body equations, which are then discretized using the default integrator settings (Euler integration, $\Delta t = 0.008$).

The control constraints are box constraints within $[-1, 1]^3$ taken from the default settings.
The state constraints are represented as the set $h(\vx) = \max(h_0(\vx), h_1(\vx)) \leq 0$, where 
\begin{equation}
    h_0(\vx) \coloneqq 0.7 - p_z, \quad
    h_1(\vx) \coloneqq \abs{\theta} - 0.2,
\end{equation}
which represent maintaining a minimum height and preventing the torso from tipping over too much, and are taken from the default settings.
The goal set $\GoalSet$ is defined as the set of states where $p_x$ is within the set $[2.8, 3.0]$. This is represented via the cost function
\begin{equation}
    l(\vx) \coloneqq [\abs{p_x - 2.9} - 0.1]^+.
\end{equation}

\newpage
\subsection{F16 Ground Collision Avoidance in a Low Altitude Flight Corridor}
\begin{table}[H]
\centering
\begin{booktabs}{
  colspec = {rllrll},
  cell{1}{1,4} = {c=3}{c}, }
\toprule 
    States & _ & _ & Controls & _ & _ \\ \cmidrule[lr]{1-3} \cmidrule[lr]{4-6}
    Index & Symbol & Description & Index & Symbol & Description \\ \midrule
     0 & $v_T$         & Air speed                          & 0 & $Nz_d$     & Setpoint for accleration                     \\
     1 & $\alpha$      & Angle of attack                    & 1 & $Ps_d$     & Setpoint for stability roll rate             \\
     2 & $\beta$       & Angle of sideslip                  & 2 & $NypR_d$   & Setpoint for side acceleration and yaw rate  \\
     3 & $\phi$        & Roll                               & 3 & $\delta_t$ & Throttle                                     \\
     4 & $\theta$      & Pitch                              &   &            &                                              \\
     5 & $\psi$        & Yaw                                &   &            &                                              \\
     6 & $P$           & Roll rate                          &   &            &                                              \\
     7 & $Q$           & Pitch rate                         &   &            &                                              \\
     8 & $R$           & Yaw rate                           &   &            &                                              \\
     9 & $p_N$         & Northward displacement             &   &            &                                              \\
    10 & $p_E$         & Eastward displacement              &   &            &                                              \\
    11 & $p_U$         & Altitude                           &   &            &                                              \\
    12 & $\text{pow}$  & Engine power lag                   &   &            &                                              \\
    13 & $Nz$          & Upward acceleration                &   &            &                                              \\
    14 & $Ps$          & Stability roll rate                &   &            &                                              \\
    15 & $NypR$        & Side acceleration and yaw rate     &   &            &                                              \\
    15 & $\mathcal{V}$ & Valid mask                         &   &            &                                              \\
    \bottomrule
\end{booktabs}
\caption{States and Controls for the F16}
\label{tab:f16:overview}
\end{table}

The F16 is a system with $17$ state and $4$ control dimensions (see \Cref{tab:f16:overview}) based on \cite{heidlauf2018verification}.
Note that the original system includes only $16$ states.
We have added the final state $\mathcal{V}$ to prevent the state from exiting the region where the F16 model is numerically accurate.
The continuous-time dynamics are defined in \cite{heidlauf2018verification} is a standard model used in aerospace engineering and described extensively in the textbook by \citet{stevens2004aircraft}. Notably, the dynamics makes use of look-up tables for aspects such as the engine model and aerodynamic coefficients.
The discrete-time dynamics are defined by integrating the continuous-time dynamics using the RK4 integrator with a step size of $\Delta t = 0.05$.
Moreover, when the system exits the region where the model is valid, defined as the set
\begin{equation}
    \Set{\vx |
        \alpha \in [\underline{\alpha}, \overline{\alpha}], \beta \in [\underline{\beta}, \overline{\beta}], \theta \in [-\frac{\pi}{2}, \frac{\pi}{2}]
    },
\end{equation}
we set $\mathcal{V}$ to $0$ and stop integrating the dynamics to prevent the model from misbehaving.
The bounds for $\alpha$ and $\beta$ are taken as the limits of the aerodynamic data tables, while the bounds for the pitch $\theta$ are used to avoid the singularity due to the use of Euler angles.

The control constraints are box constraints defined as
\begin{equation}
    Nz_d \in [-10, 15], \quad
    Ps_d \in [-10, 10], \quad
    NypR_d \in [-10, 10], \quad
    \delta_t \in [0, 1]
\end{equation}
The state constraints are represented as the set $h(\vx) = \max_i h_i(\vx) \leq 0$, where the constraint functions $h_i$ are defined as
\begin{subequations}
\begin{align}
   \text{(Avoid ground and stay below ceiling)} && h_0(\vx) &\coloneqq \max(0 - p_U, p_U - 1000) \, / \, 200, \\
   \text{(Keep $\alpha$ valid)} && h_1(\vx) &\coloneqq \max(\underline{\alpha} - \alpha, \alpha - \overline{\alpha}) \, / \, 0.2, \\
   \text{(Keep $\beta$ valid)} && h_2(\vx) &\coloneqq \max(\underline{\beta} - \beta, \beta - \overline{\beta}) \, / \, 0.2, \\
   \text{(Keep $\theta$ valid)} && h_3(\vx) &\coloneqq \max(\underline{\theta} - \theta, \theta - \overline{\theta}) \, / \, 0.2, \\
   \text{(Stay within the flight corridor)} && h_4(\vx) &\coloneqq \max(-200 - p_E, p_E - 200) \, / \, 50.
\end{align}
\end{subequations}
The goal region $\GoalSet$ is defined as the set of states where the altitude $p_U$ is within the set $[50, 150]$. Note that this set is very close to the ground and thus the reason why we call this task ``low-altitude flight corridor''. We implement this via the cost function
\begin{equation}
    l(\vx) \coloneqq \max \big(50 - p_U, p_U - 150 \big) \,/\, 250.
\end{equation}

\end{appendices}

\end{document}